\documentclass{article}

\usepackage[dblblindworkshop,final]{neurips_2026}

\usepackage[utf8]{inputenc}
\usepackage[T1]{fontenc}
\usepackage{hyperref}
\usepackage{url}
\usepackage{booktabs}
\usepackage{placeins}
\usepackage{float}
\usepackage{amsfonts}
\usepackage{amsmath}
\usepackage{amssymb}
\usepackage{nicefrac}
\usepackage{microtype}
\usepackage{xcolor}
\usepackage{graphicx}
\usepackage{array}
\usepackage{multirow}
\usepackage{subcaption}
\usepackage{xspace}

\newcolumntype{N}{>{\centering\arraybackslash}p{0.92cm}}

\graphicspath{{figures/}}

\usepackage[font=small,labelfont=bf]{caption}
\usepackage{enumitem}
\setlist{itemsep=2pt,topsep=4pt,parsep=0pt,partopsep=0pt}
\usepackage{titlesec}
\titlespacing*{\section}{0pt}{7pt plus 2pt minus 2pt}{4pt}
\titlespacing*{\subsection}{0pt}{6pt plus 2pt minus 2pt}{3pt}
\titlespacing*{\subsubsection}{0pt}{5pt plus 2pt minus 2pt}{2pt}
\titlespacing*{\paragraph}{0pt}{4pt plus 2pt minus 1pt}{6pt}

\usepackage{kotex}

\newif\ifcameraready
\camerareadytrue

\newcommand{\dataset}{ForestBelongings\xspace}

\workshoptitle{Grounded and Faithful Vision-Language Models for Real-World Deployment}

\title{Query-Composition Sensitivity of Open-Vocabulary Detectors in Under-Canopy Search and Rescue}
\hypersetup{pdftitle={Query-Composition Sensitivity of Open-Vocabulary Detectors in Under-Canopy Search and Rescue}}

\author{%
  Deokyun Kim\thanks{Equal contribution.} \quad
  Giyoung Lee\footnotemark[1] \quad
  Yookyung Kim \quad
  Myungseok Ki \quad
  Jihun Cha \\[2pt]
  Autonomous UAV Research Section \\
  Electronics and Telecommunications Research Institute \\
  \texttt{\{deokyunkim, giyoung, yk.kim, serdong, jihun\}@etri.re.kr} \\
}

\begin{document}

\maketitle

\begin{abstract}
In under-canopy search and rescue, visual evidence includes the missing person and personal belongings likely lost or left behind as critical search clues. Such belongings span categories too diverse to define exhaustively in advance and are highly variable in appearance, motivating open-vocabulary detectors (OVDs) rather than fixed closed-set detectors. However, in operational search settings, operators often construct text queries that name different combinations of search targets depending on their search intent. For example, operators often query a person alone or together with clothing or other belongings, and naming additional targets is likely to affect the predictions even for a search target named in every query. In this work, we investigate how text-query composition affects OVD predictions and detection performance in search and rescue scenarios. We adapt OVDs to under-canopy search and evaluate the adapted OVDs using a controlled text-query protocol that varies which person and belongings targets are named together and how each target is named. Experimental results show that predictions and detection performance for the same search target can differ depending on the other targets named in the query for fusion-based OVDs, while the alignment-based detector remains invariant. Adaptation also transfers unevenly across operationally equivalent target names. These findings demonstrate that query construction should be explicitly considered when evaluating and deploying OVDs. To support this study, we introduce \dataset, a benchmark for detecting personal belongings in forest environments, containing 31,897 under-canopy images with 39,831 annotations, publicly available at \url{https://huggingface.co/datasets/etri/ForestBelongings}.
\end{abstract}

\section{Introduction}
\label{sec:intro}

Micro aerial vehicles (MAVs) provide a practical platform for forest search and
rescue since they can collect under-canopy visual evidence in terrain that is
difficult for ground teams to access
\citep{laina2024,hong2024,ren2025,jarinlipschitz2022,zhou2021fuel,lin2022,kwon2024}.
Detecting missing persons in this setting is challenging: they are small,
partially occluded, and embedded in cluttered forest backgrounds
\citep{forestpersons}. The relevant evidence is also not limited to the missing
person. Detached personal belongings in the \emph{clothes}, \emph{hat},
\emph{shoes}, \emph{gloves}, and \emph{bag} categories provide important
search evidence, but they are highly variable in appearance, forest-floor
context, and even in the most appropriate name for the same object. This makes
open-vocabulary detection (OVD) attractive for forest search, since the detector
can be asked for
scenario-dependent visual evidence beyond a fixed closed-set person category
\citep{li2022glip,minderer2023owlv2,liu2024groundingdino}.

\begin{figure}[tb]
  \centering
  \includegraphics[width=1.0\linewidth]{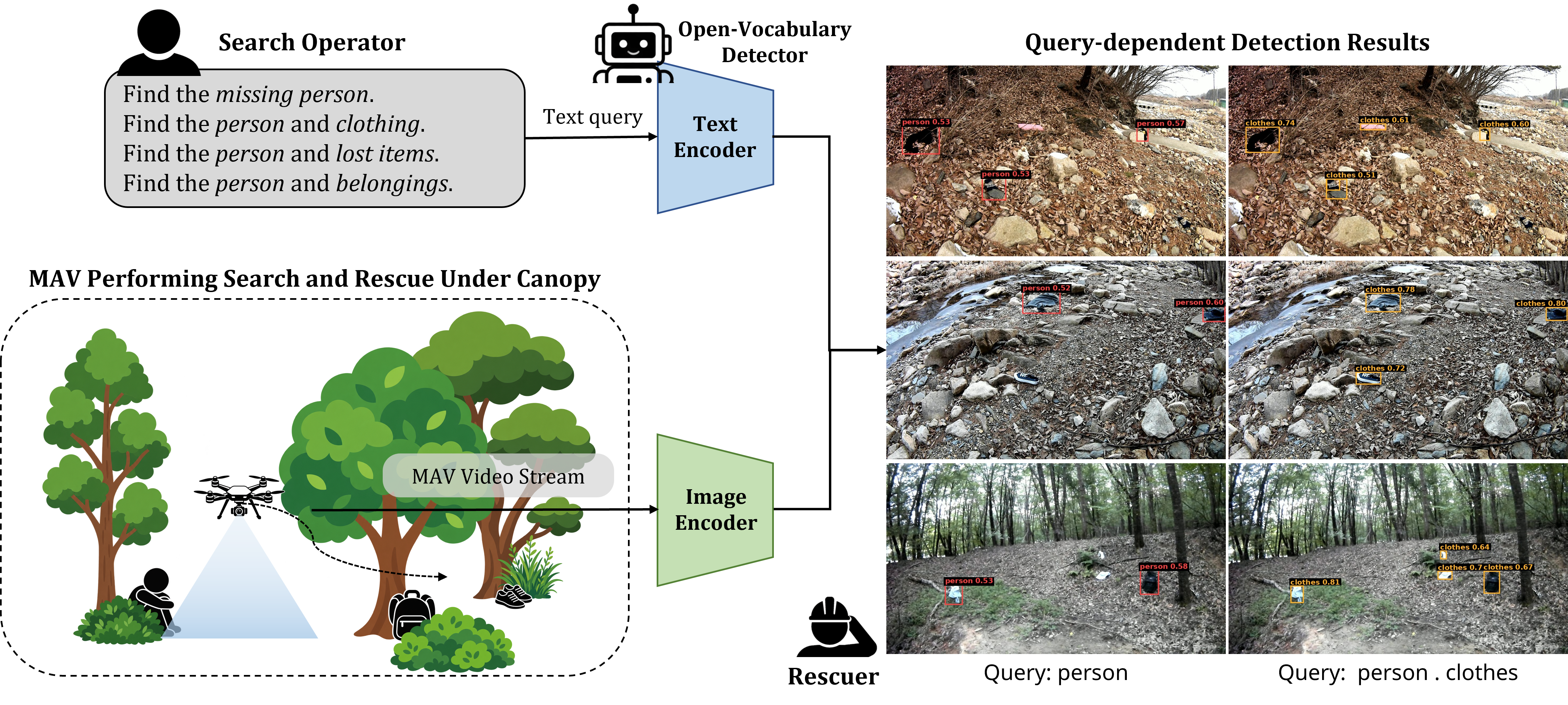}
  \caption{\textbf{Overview of OVD-based under-canopy search.} The same
  imagery captured in an under-canopy environment is evaluated under
  different text-query compositions, yielding
  different detections for the shared \emph{person} target. In
  deployment, such query-dependent outcomes can affect the decision to
  dispatch rescuers.}
  \label{fig:overview}
\end{figure}

However, this text-query flexibility has consequences beyond evaluation: query
composition can alter measured detection performance and, in a search and
rescue setting, is likely to affect whether the missing person is found and
rescued.
In practice, operators often query for a person alone, all relevant belongings,
or a specific clue such as clothes
(Figure~\ref{fig:overview}). If a person detection is lost or reassigned as a
belonging under a different query composition, the corresponding location risks
being deprioritized, potentially reducing the likelihood of timely rescue. These are
all reasonable queries for the same detection task. However, most OVD methods
jointly encode the queried categories before scoring candidate boxes. Consequently, the
average precision (AP) for a category such as \emph{person} is sensitive to the other named categories
included in the query. Evaluation with a single fixed query vocabulary, such as
\emph{person} with all belongings categories, does not fully
capture detector behavior under practical query-formulation changes. From a
deployment perspective, this is a grounding-reliability problem: the visual
evidence for the shared target is unchanged, yet its localization can depend on
which additional categories the operator names. Query composition should
therefore be treated as an evaluation variable rather than as a fixed component
of the experimental setup.

We study query-composition sensitivity under an assumed deployment setting: an
OVD used for under-canopy search and rescue (SAR), where the query is formulated by a
search operator at mission time rather than fixed in advance as part of the
evaluation protocol. To the best of our knowledge, this is the first study to
systematically evaluate query-composition sensitivity of OVDs under an assumed
SAR deployment scenario. Prior work on query sensitivity has
focused on generic benchmarks or other application domains
\citep{miller2024opensetvlm,lundqvist2026botanical}, whereas existing
forest-search benchmarks evaluate person detection using a single fixed
vocabulary. We therefore combine ForestPersons \citep{forestpersons} with
\dataset, an under-canopy dataset of detached personal belongings collected for
object detection. The resulting benchmark combines person-present forest scenes
with \dataset scenes that contain detached belongings or no annotated search
target, and evaluates them under a controlled query-composition protocol. Our
primary comparisons manipulate both query size
and the identity of the belonging category included alongside \emph{person},
allowing us to assess whether detection performance for a fixed target category
remains stable as the operator changes the accompanying clue category.

\paragraph{Contributions.}
\begin{itemize}
  \item We introduce \dataset, a dataset of 31{,}897
        under-canopy images of detached personal belongings, with 39{,}831
        bounding-box annotations. To the best of our knowledge, \dataset is
        the first public benchmark for
        detecting personal belongings as visual cues in missing-person search
        and rescue in forest environments.
  \item We quantify query-composition sensitivity in OVD for forest search
        by measuring changes in AP and average recall (AR) across
        different query compositions before and after domain adaptation,
        thereby assessing whether adaptation improves robustness to query
        formulation.
  \item We further establish a controlled evaluation protocol with query
        types reflecting plausible SAR mission formulations:
        a single category, a \emph{person} with one clue, and a
        \emph{person} with all clues. This design
        disentangles the effect of query-set size from that of clue identity,
        while complementing aggregate AP and AR with instance-level
        consistency metrics.
\end{itemize}

\section{Related Work}
\label{sec:related}

OVDs use language-conditioned recognition interfaces
that allow object categories to be specified at inference time
\citep{li2022glip,minderer2023owlv2,liu2024groundingdino,omdet_turbo}. Prior
work has shown that detection performance depends on how this text interface is
constructed, including prompt representations, richer descriptions, and
class-aware prompt refinement
\citep{du2022detpro,li2023desco,bianchi2024fgovd,fernandez2026cgap}. More recent
studies further show that the query set itself matters: increasing query-set
size can degrade open-vocabulary recognition \citep{miller2024opensetvlm}.
These studies primarily
examine query wording, prompt design, or query-set size. Broader robustness
studies similarly show that model behavior can depend on input or prompt
formulation even when the underlying task remains unchanged
\citep{sclar2024format,yuksekgonul2023aro,thrush2022winoground,hsieh2023sugarcrepe}.
Together, these studies establish that OVD predictions are sensitive to how
text queries are constructed. Building on this perspective, we examine
query-composition sensitivity for a fixed target in under-canopy search and
rescue.

Recent OVD research has increasingly considered deployment-oriented and
domain-specific settings, including agricultural detection, remote sensing,
and aerial perception
\citep{lundqvist2026botanical,pan2025lae,tao2025uavovd}. Under-canopy search presents a distinct
challenge, where targets must be localized near the forest floor under severe
occlusion, camouflage, and visual clutter. ForestPersons provides a benchmark
for missing-person detection in this environment \citep{forestpersons}, but
focuses on person detection under a fixed target vocabulary. Missing-person
search, however, depends not only on direct person sightings but also on
detached belongings that can serve as critical clues for directing the
subsequent search. We therefore extend this setting to person-and-belongings
detection under operationally relevant object queries. Accordingly, we treat
query composition as part of the assumed deployment scenario rather than merely
as a linguistic property of the input.

\section{Forest Search Benchmark and Query-Composition Evaluation Protocol}
\label{sec:method}

\subsection{Preliminaries: OVD Query Processing and Domain Adaptation}
\label{sec:method:preliminaries}

OVDs take an image and a text query as joint
inputs and localize the search targets named in the query. OVD detections
are therefore sensitive to \emph{query construction}, including the number of
named search targets and the categories named alongside a shared target. Query
construction differs across SAR scenarios.
To leverage the vision-language knowledge encoded in pretrained OVDs during
adaptation to under-canopy imagery, we apply LoRA \citep{lora} to four
representative detectors: GroundingDINO \citep{liu2024groundingdino}, GLIP
\citep{li2022glip}, OmDet-Turbo \citep{omdet_turbo}, and OWLv2
\citep{minderer2023owlv2}. Comparing the LoRA-adapted detectors with the
corresponding zero-shot detectors allows us to assess how domain adaptation
affects AP and AR as well as query-composition robustness.

The OVDs support text-conditioned detection and LoRA-based domain adaptation
while keeping the pretrained image- and text-encoder weights fixed, enabling
the adapted models to utilize visual-language knowledge acquired during
large-scale pre-training.
The OVDs differ mainly in where text-query representations interact
with image features. GroundingDINO, GLIP, and OmDet-Turbo are fusion-based:
text-query representations are fused with image features before boxes are scored
\citep{li2022glip,liu2024groundingdino,omdet_turbo}. OWLv2 employs an
alignment-based architecture that computes query-specific logits from
independently encoded text queries and image patches
\citep{minderer2023owlv2}.

\subsection{ForestBelongings and Benchmark Construction}
\label{sec:method:benchmark}

\begin{figure}[tb]
  \centering
  \includegraphics[width=\linewidth]{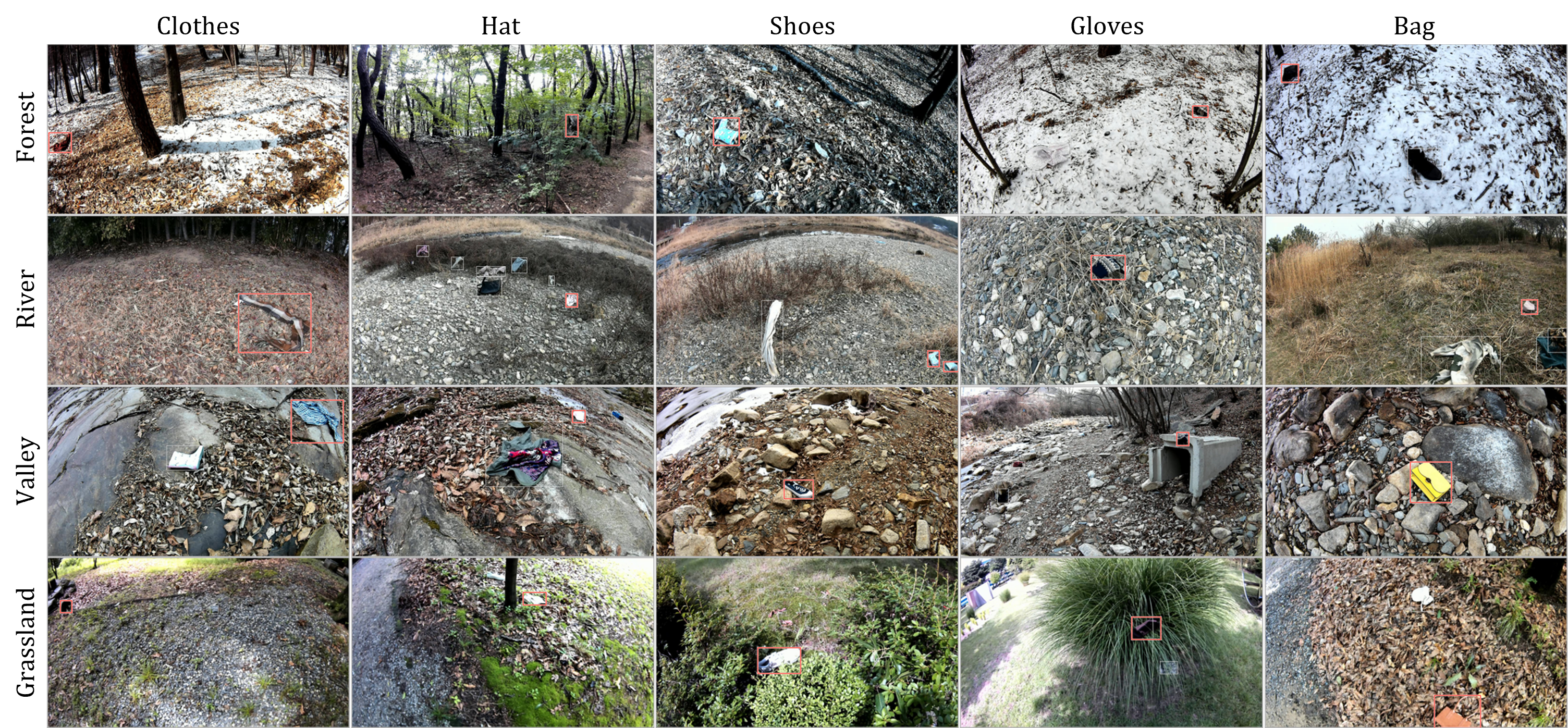}
  \caption{Qualitative overview of \dataset. Rows show collection environments,
  columns show belongings categories, and red boxes indicate annotated objects.}
  \label{fig:forestbelongings-overview}
\end{figure}

\paragraph{\dataset.}
\dataset is an under-canopy object-detection dataset for personal
belongings as clues for locating missing persons. To approximate the
under-canopy MAV viewpoints in ForestPersons \citep{forestpersons}, we placed
detached belongings in forests and captured them with a handheld camera.
The dataset contains 31{,}897 images across 250 sequences, 39{,}831 annotations,
and 3{,}423 no-object frames retained as negative samples for target-absent
under-canopy scenes. Here, \emph{category names} denote the human-defined
annotation labels. The combined
evaluation vocabulary consists of the ForestPersons category \emph{person} and
the \dataset belongings categories used in the main benchmark --- \emph{clothes}, \emph{hat},
\emph{shoes}, \emph{gloves} and \emph{bag} --- selected as the belongings
that hikers most commonly carry. Figure~\ref{fig:forestbelongings-overview}
illustrates detached belongings that may be encountered during
under-canopy searches. Incidental \emph{person} annotations in
\dataset are unintentionally captured people. Sequence-disjoint splits and
per-class counts are provided in Appendix~\ref{app:dataset}.

\paragraph{Benchmark construction with ForestPersons.}
To train OVDs for under-canopy SAR and evaluate detection
reliability under operationally relevant conditions, we combine \dataset with
ForestPersons: ForestPersons
contributes RGB forest imagery annotated for \emph{person}, and \dataset
contributes detached belongings that can serve as operational clues. Together, these test splits
form the \emph{Under-Canopy Person-and-Belongings Detection Benchmark}, our
main evaluation set. It contains 14{,}980 RGB images: 10{,}553 ForestPersons images
and 4{,}427 \dataset images, with 20{,}895 person boxes and 5{,}723 benchmark
belongings boxes.

\paragraph{Belongings as search clues.}
We define a belonging as a portable object commonly carried by hikers that
is \emph{spatially separated from any body}. Clothing worn by a
person, a bag on a shoulder, or shoes on the feet remain part of the
person detection problem and receive no separate belongings annotation. In
\dataset, a jacket on the forest floor, a detached bag, or shoes away from a
body are annotated as independent clue objects. This distinction is
operationally important: locating a missing person ends the search and
initiates rescue, whereas locating a detached belonging provides a clue
that directs further search to the surrounding area.

\subsection{Query-Composition Evaluation Protocol and Metrics}
\label{sec:method:queries}
\label{sec:method:metrics}

In under-canopy SAR, OVD query composition reflects whether the
operational objective is to find the missing person or detached belongings as
potential clues or both.
To isolate the effect of query composition, we vary which belongings
categories are named alongside \emph{person} and measure whether the
\emph{person} prediction is preserved. The protocol is applied to both
zero-shot and fine-tuned OVDs to assess AP, AR, and query-composition
robustness for under-canopy SAR.

The protocol is organized around three query types: (1) singleton queries
ask for a single category at a time; (2) the full-category query
asks jointly for \emph{person} and the belongings vocabulary, providing the
person-and-belongings operating point; and (3) companion-query variants
pair \emph{person} with a single belongings category or with the belongings
vocabulary, allowing us to contrast individual companion categories with the
full-category query.

Aggregate AP and AR can remain similar across queries even when instance-level
detection outcomes are inconsistent. In particular, a person detected under one
query may be missed under another. To capture such instance-level inconsistencies, which are
critical in SAR, we further measure instance-level
prediction consistency across query compositions.

\section{Experiments}
\label{sec:exp}

\begin{table}[!t]
  \caption{Class-wise AP and AR for \textbf{pretrained zero-shot} OVDs under the full-category query.}
  \label{tab:zeroshot-forest}
  \centering
  \footnotesize
  \setlength{\tabcolsep}{3pt}
  \begin{tabular*}{\textwidth}{@{\extracolsep{\fill}}l cc cc cc cc@{}}
    \toprule
    & \multicolumn{2}{c}{GroundingDINO} & \multicolumn{2}{c}{GLIP}
      & \multicolumn{2}{c}{OmDet-Turbo} & \multicolumn{2}{c}{OWLv2} \\
    \cmidrule(lr){2-3} \cmidrule(lr){4-5}
    \cmidrule(lr){6-7} \cmidrule(lr){8-9}
    Class & AP & AR & AP & AR & AP & AR & AP & AR \\
    \midrule
    \texttt{person} & 57.80 & 71.36 & 56.79 & 70.23 & 54.29 & 65.93 & 58.00 & 74.39 \\
    \cmidrule(lr){1-9}
    \texttt{clothes} & \phantom{0}4.12 & 78.90 & 11.97 & 71.72 & 33.61 & 72.55 & 24.37 & 79.00 \\
    \texttt{hat} & \phantom{0}0.84 & 85.40 & \phantom{0}1.73 & 79.42 & \phantom{0}1.06 & 72.92 & \phantom{0}7.18 & 82.18 \\
    \texttt{shoes} & \phantom{0}6.78 & 78.59 & \phantom{0}1.96 & 66.85 & \phantom{0}9.57 & 71.29 & 24.22 & 81.97 \\
    \texttt{gloves} & \phantom{0}0.55 & 76.53 & \phantom{0}1.59 & 67.53 & \phantom{0}1.03 & 69.14 & \phantom{0}6.54 & 77.95 \\
    \texttt{bag} & 20.06 & 74.54 & \phantom{0}5.12 & 72.34 & 12.35 & 68.25 & 25.80 & 77.75 \\
    \midrule
    \emph{Belongings} & \emph{\phantom{0}6.47} & \emph{78.79} & \emph{\phantom{0}4.47} & \emph{71.57} & \emph{11.52} & \emph{70.83} & \emph{17.62} & \emph{79.77} \\
    \emph{Mean} & \emph{15.02} & \emph{77.55} & \emph{13.19} & \emph{71.35} & \emph{18.65} & \emph{70.01} & \emph{24.35} & \emph{78.87} \\
    \bottomrule
  \end{tabular*}

\par\vspace{4pt}
  \caption{\emph{Person} AP and AR for \textbf{pretrained zero-shot} OVDs under
  person-query variants. \emph{Person range} is max--min within each metric over
  the six companion-query variants, including the full-category query.}
  \label{tab:pair-identity}
  \centering
  \footnotesize
  \setlength{\tabcolsep}{2.2pt}
  \begin{tabular*}{\textwidth}{@{\extracolsep{\fill}}l cc cc cc cc@{}}
    \toprule
    & \multicolumn{2}{c}{GroundingDINO}
    & \multicolumn{2}{c}{GLIP}
    & \multicolumn{2}{c}{OmDet-Turbo}
    & \multicolumn{2}{c}{OWLv2} \\
    \cmidrule(lr){2-3} \cmidrule(lr){4-5}
    \cmidrule(lr){6-7} \cmidrule(lr){8-9}
    Text query & AP & AR & AP & AR & AP & AR & AP & AR \\
    \midrule
    \texttt{person} \emph{(base)}
      & 56.49 & 71.57 & 48.64 & 67.16 & 53.19 & 66.10 & 58.00 & 74.41 \\
    \midrule
    \texttt{person clothes}
      & 58.73 & 73.57 & 57.13 & 70.66 & 52.50 & 66.16 & 58.00 & 74.41 \\
    \texttt{person hat}
      & 55.51 & 70.38 & 57.50 & 70.99 & 51.41 & 65.85 & 58.00 & 74.41 \\
    \texttt{person shoes}
      & 56.49 & 70.99 & 58.29 & 71.30 & 53.87 & 65.54 & 58.00 & 74.41 \\
    \texttt{person gloves}
      & 56.92 & 70.99 & 58.74 & 71.49 & 52.23 & 65.92 & 58.00 & 74.41 \\
    \texttt{person bag}
      & 55.20 & 71.07 & 55.84 & 69.95 & 53.24 & 65.93 & 58.00 & 74.41 \\
    \midrule
    \emph{full-category query}
      & 57.80 & 71.36 & 56.79 & 70.23 & 54.29 & 65.93 & 58.00 & 74.39 \\
    \midrule
    \emph{Person range (max--min)}
      & \emph{3.53} & \emph{3.19}
      & \emph{2.90} & \emph{1.54}
      & \emph{2.88} & \emph{0.63}
      & \emph{0.00} & \emph{0.02} \\
    \bottomrule
  \end{tabular*}
\end{table}

\subsection{Setup}
\label{sec:exp:setup}

For benchmark adaptation, each training image is paired with the
\emph{full-category query}, \texttt{person . clothes . hat . shoes . gloves .
bag .}, which lists all target categories in a fixed order. Each detector
is adapted using the model-native detection objective and label-assignment
scheme, preserving the training formulation of the original model.
Across detectors, adaptation uses LoRA
\citep{lora} with rank $r=16$, scaling parameter $\alpha=32$, dropout $0.05$,
and no bias parameters. Training is performed for one epoch with
an effective batch size of $16$ using AdamW, bf16 precision, gradient
clipping at $1.0$, and a cosine learning-rate schedule with $3\%$ warmup.
LoRA parameters are optimized with a learning rate of $10^{-4}$, while
trainable detection heads use $10^{-5}$. Each adaptation is repeated
with three random seeds on a single NVIDIA RTX 6000 Ada GPU.

For evaluation, detector performance is measured across query compositions
defined by the categories named together in each text query. In the result
tables, \texttt{person clothes} means that
\emph{clothes} is queried in addition to \emph{person}. Every companion-query
variant therefore retains \emph{person} as the shared target and adds either
one belongings category or the belongings vocabulary. All queries are
evaluated on the full pooled test set. AP and AR are computed over IoU
thresholds $.50{:}.95$ following
the COCO evaluation protocol, using pycocotools \citep{coco}, with a score
threshold of $0.0$ and COCO maxDets $=100$. Fine-tuned results are reported as
the mean and sample standard deviation across three runs. Additional details
on the evaluation protocol, model-native post-processing, trainable components,
and model-specific adaptation costs are provided in
Appendix~\ref{app:setup-details}.

\subsection{Evaluation Protocol}
\label{sec:exp:protocol}

All experiments follow the query-composition protocol in
Section~\ref{sec:method:queries}. The full-category query establishes
the joint person-and-belongings operating point. Companion-query variants test
whether \emph{person} predictions change when a single clue category is added
or when the belongings vocabulary is named, and the same contrasts are
repeated after LoRA domain adaptation. Beyond per-query AP and AR, we evaluate
the consistency of \emph{person} detections across query compositions. Let
$q_0$ denote the \texttt{person} reference query and $q$ a companion-query
variant. A ground-truth person is detected under a query when at least one
same-class prediction matches it at IoU $\geq .50$ after the fixed
post-processing used for evaluation. Following the concepts of negative and
positive flips for measuring prediction consistency \citep{yan2021positive},
the negative flip rate (NFR) is the fraction of persons detected under $q_0$
but lost under $q$, whereas the positive flip rate (PFR) is the fraction of
persons missed under $q_0$ but newly detected under $q$. We also report the
stealing rate (SR), the fraction of
ground-truth persons whose \emph{person} detection is lost under the variant
query but whose location remains covered by a belongings prediction. For
person-absent subsets, we additionally count person false positives and whether
they overlap annotated belongings (Appendix~\ref{app:person-absent}).

\begin{figure}[t]
  \centering
  \includegraphics[width=1.0\linewidth]{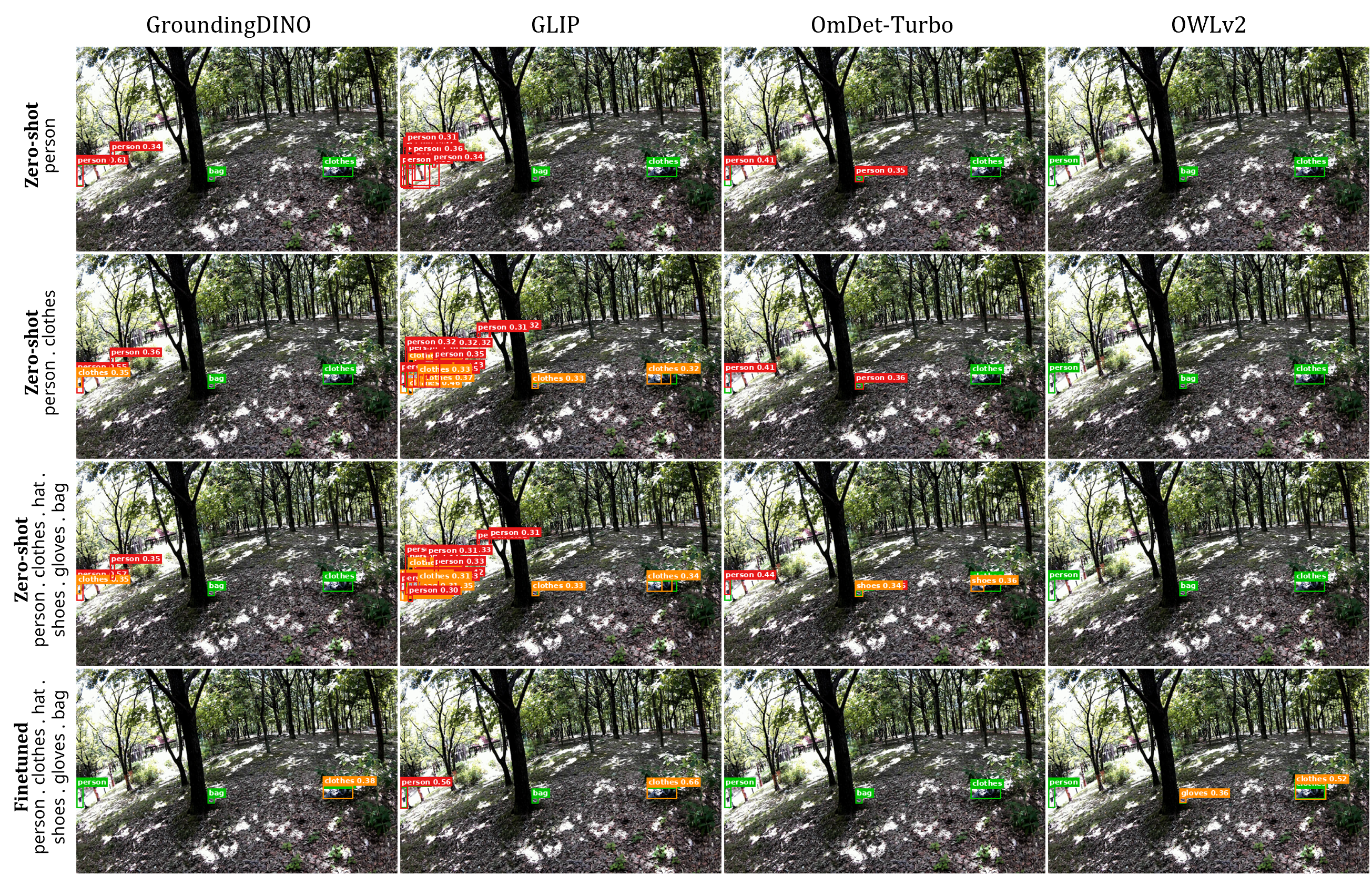}
  \caption{Qualitative results for the four OVDs on the \dataset test split
  under the evaluated query settings (green: ground truth; red:
  \emph{person}; orange: belongings).}
  \label{fig:query-grid}
\end{figure}

\subsection{Analysis on the Under-Canopy Person-and-Belongings Detection Benchmark}
\label{sec:exp:analysis}

\subsubsection{Zero-shot OVD performance on the benchmark}
\label{sec:exp:singletons}
\label{sec:exp:zeroshot-query}

Table~\ref{tab:zeroshot-forest} reports zero-shot OVD
performance on the Under-Canopy Person-and-Belongings Detection
Benchmark described in Section~\ref{sec:method:benchmark} under the
full-category query. The \emph{person} category is detected
reasonably well across models, with AP ranging from $54.29$ for OmDet-Turbo to
$58.00$ for OWLv2. In contrast,
belongings AP remains much lower despite high AR, indicating that zero-shot OVD
can produce candidate boxes for these objects but struggles with reliable
under-canopy belongings detection.
Given these zero-shot results, we adopt \emph{person} as the shared target
of the query-composition analysis, since in deployment the detection of
the missing person is the decisive outcome.

Table~\ref{tab:pair-identity} shows that, for every detector except
OWLv2, \emph{person} AP and AR depend on which belongings category is
requested alongside, although the \emph{person} target itself is unchanged
across queries. Across the companion-query variants, zero-shot \emph{person} AP
spans $3.53$ for GroundingDINO, $2.90$ for GLIP, and $2.88$ for
OmDet-Turbo. In contrast, OWLv2 yields identical \emph{person} AP and only a
$0.02$-point AR range across queries, consistent with its alignment-based
architecture that computes query-specific logits from independently encoded
text queries and image patches. Relative to the \texttt{person}-only query, \emph{person} AP rises
for GLIP under every paired query and for GroundingDINO when
\emph{clothes} is requested alongside, whereas for OmDet-Turbo most paired
queries reduce \emph{person} AP below its \texttt{person}-only value. These results suggest that, for
OVDs operated in SAR, the likelihood that the missing person
is detected depends on the text-query composition. Per-category belongings
results are in Appendix~\ref{app:singletons}.

\begin{table}[!t]
  \caption{Class-wise AP and AR for \textbf{LoRA fine-tuned} OVDs under the
  full-category query, as the mean over three adaptation seeds. The corresponding
  \textbf{pretrained zero-shot} values are in Table~\ref{tab:zeroshot-forest}.}
  \label{tab:zs-vs-ft}
  \centering
  \footnotesize
  \setlength{\tabcolsep}{1.6pt}
  \begin{tabular*}{\textwidth}{@{\extracolsep{\fill}}l cc cc cc cc@{}}
    \toprule
    & \multicolumn{2}{c}{GroundingDINO} & \multicolumn{2}{c}{GLIP}
      & \multicolumn{2}{c}{OmDet-Turbo} & \multicolumn{2}{c}{OWLv2} \\
    \cmidrule(lr){2-3} \cmidrule(lr){4-5}
    \cmidrule(lr){6-7} \cmidrule(lr){8-9}
    Class & AP & AR & AP & AR & AP & AR & AP & AR \\
    \midrule
    \texttt{person} & 64.30{\tiny\,$\pm$\,0.24} & 78.14{\tiny\,$\pm$\,0.19} & 67.93{\tiny\,$\pm$\,0.24} & 76.52{\tiny\,$\pm$\,0.12} & 55.26{\tiny\,$\pm$\,0.37} & 66.75{\tiny\,$\pm$\,0.16} & 59.00{\tiny\,$\pm$\,3.17} & 73.22{\tiny\,$\pm$\,1.02} \\
    \cmidrule(lr){1-9}
    \texttt{clothes} & 68.97{\tiny\,$\pm$\,0.29} & 83.67{\tiny\,$\pm$\,0.13} & 70.26{\tiny\,$\pm$\,0.14} & 83.21{\tiny\,$\pm$\,0.06} & 56.02{\tiny\,$\pm$\,0.28} & 74.70{\tiny\,$\pm$\,0.15} & 64.74{\tiny\,$\pm$\,1.35} & 79.45{\tiny\,$\pm$\,0.77} \\
    \texttt{hat} & 53.26{\tiny\,$\pm$\,0.76} & 85.72{\tiny\,$\pm$\,0.68} & 56.49{\tiny\,$\pm$\,1.25} & 84.09{\tiny\,$\pm$\,0.61} & 37.33{\tiny\,$\pm$\,1.31} & 74.99{\tiny\,$\pm$\,0.39} & 58.28{\tiny\,$\pm$\,1.91} & 82.41{\tiny\,$\pm$\,0.55} \\
    \texttt{shoes} & 54.25{\tiny\,$\pm$\,0.08} & 84.20{\tiny\,$\pm$\,0.07} & 54.38{\tiny\,$\pm$\,1.80} & 83.87{\tiny\,$\pm$\,0.54} & 35.53{\tiny\,$\pm$\,1.44} & 73.66{\tiny\,$\pm$\,0.41} & 57.84{\tiny\,$\pm$\,1.86} & 80.63{\tiny\,$\pm$\,0.52} \\
    \texttt{gloves} & 34.29{\tiny\,$\pm$\,1.36} & 79.33{\tiny\,$\pm$\,0.64} & 29.72{\tiny\,$\pm$\,3.20} & 78.50{\tiny\,$\pm$\,0.64} & 21.89{\tiny\,$\pm$\,1.45} & 72.10{\tiny\,$\pm$\,0.36} & 37.87{\tiny\,$\pm$\,1.52} & 77.42{\tiny\,$\pm$\,1.84} \\
    \texttt{bag} & 48.84{\tiny\,$\pm$\,1.80} & 79.71{\tiny\,$\pm$\,0.22} & 35.76{\tiny\,$\pm$\,1.85} & 76.09{\tiny\,$\pm$\,0.83} & 33.51{\tiny\,$\pm$\,0.45} & 68.02{\tiny\,$\pm$\,0.24} & 54.27{\tiny\,$\pm$\,0.32} & 76.40{\tiny\,$\pm$\,0.69} \\
    \midrule
    \emph{Belongings} & \emph{51.92{\tiny\,$\pm$\,0.45}} & \emph{82.53{\tiny\,$\pm$\,0.02}} & \emph{49.32{\tiny\,$\pm$\,0.46}} & \emph{81.15{\tiny\,$\pm$\,0.35}} & \emph{36.85{\tiny\,$\pm$\,0.40}} & \emph{72.70{\tiny\,$\pm$\,0.04}} & \emph{54.60{\tiny\,$\pm$\,1.24}} & \emph{79.26{\tiny\,$\pm$\,0.50}} \\
    \emph{Mean} & \emph{53.98{\tiny\,$\pm$\,0.41}} & \emph{81.80{\tiny\,$\pm$\,0.05}} & \emph{52.43{\tiny\,$\pm$\,0.41}} & \emph{80.38{\tiny\,$\pm$\,0.30}} & \emph{39.92{\tiny\,$\pm$\,0.37}} & \emph{71.70{\tiny\,$\pm$\,0.03}} & \emph{55.33{\tiny\,$\pm$\,1.56}} & \emph{78.26{\tiny\,$\pm$\,0.59}} \\
    \bottomrule
  \end{tabular*}

\par\vspace{4pt}
  \caption{\emph{Person} AP and AR for \textbf{LoRA fine-tuned} OVDs under
  person-query variants, as the mean over three adaptation seeds; the
  zero-shot counterparts are in Table~\ref{tab:pair-identity}. \emph{Person range} is computed within each seed.}
  \label{tab:finetuned-ap-change}
  \centering
  \footnotesize
  \setlength{\tabcolsep}{1.6pt}
  \begin{tabular*}{\textwidth}{@{\extracolsep{\fill}}l cc cc cc cc@{}}
    \toprule
    & \multicolumn{2}{c}{GroundingDINO}
    & \multicolumn{2}{c}{GLIP}
    & \multicolumn{2}{c}{OmDet-Turbo}
    & \multicolumn{2}{c}{OWLv2} \\
    \cmidrule(lr){2-3} \cmidrule(lr){4-5}
    \cmidrule(lr){6-7} \cmidrule(lr){8-9}
    Text query & AP & AR & AP & AR & AP & AR & AP & AR \\
    \midrule
    \texttt{person} \emph{(base)} & 64.73{\tiny\,$\pm$\,0.19} & 77.41{\tiny\,$\pm$\,0.27} & 50.24{\tiny\,$\pm$\,1.04} & 62.08{\tiny\,$\pm$\,1.33} & 54.69{\tiny\,$\pm$\,0.35} & 66.54{\tiny\,$\pm$\,0.04} & 59.01{\tiny\,$\pm$\,3.17} & 73.23{\tiny\,$\pm$\,1.01} \\
    \midrule
    \texttt{person clothes} & 63.02{\tiny\,$\pm$\,0.78} & 77.65{\tiny\,$\pm$\,0.12} & 67.27{\tiny\,$\pm$\,0.45} & 76.24{\tiny\,$\pm$\,0.22} & 54.98{\tiny\,$\pm$\,0.19} & 66.79{\tiny\,$\pm$\,0.11} & 59.01{\tiny\,$\pm$\,3.17} & 73.23{\tiny\,$\pm$\,1.01} \\
    \texttt{person hat} & 65.36{\tiny\,$\pm$\,0.10} & 77.24{\tiny\,$\pm$\,0.35} & 67.71{\tiny\,$\pm$\,0.30} & 76.73{\tiny\,$\pm$\,0.13} & 55.50{\tiny\,$\pm$\,0.37} & 67.06{\tiny\,$\pm$\,0.10} & 59.01{\tiny\,$\pm$\,3.17} & 73.23{\tiny\,$\pm$\,1.01} \\
    \texttt{person shoes} & 65.35{\tiny\,$\pm$\,0.21} & 77.32{\tiny\,$\pm$\,0.36} & 67.93{\tiny\,$\pm$\,0.53} & 76.87{\tiny\,$\pm$\,0.25} & 55.28{\tiny\,$\pm$\,0.31} & 66.65{\tiny\,$\pm$\,0.15} & 59.01{\tiny\,$\pm$\,3.17} & 73.23{\tiny\,$\pm$\,1.01} \\
    \texttt{person gloves} & 65.57{\tiny\,$\pm$\,0.17} & 77.42{\tiny\,$\pm$\,0.34} & 66.94{\tiny\,$\pm$\,0.25} & 76.20{\tiny\,$\pm$\,0.15} & 55.62{\tiny\,$\pm$\,0.30} & 66.90{\tiny\,$\pm$\,0.07} & 59.01{\tiny\,$\pm$\,3.17} & 73.23{\tiny\,$\pm$\,1.02} \\
    \texttt{person bag} & 65.46{\tiny\,$\pm$\,0.18} & 77.41{\tiny\,$\pm$\,0.30} & 67.55{\tiny\,$\pm$\,0.38} & 76.48{\tiny\,$\pm$\,0.13} & 55.67{\tiny\,$\pm$\,0.36} & 66.83{\tiny\,$\pm$\,0.11} & 59.01{\tiny\,$\pm$\,3.17} & 73.23{\tiny\,$\pm$\,1.01} \\
    \midrule
    \emph{full-category query} & 64.30{\tiny\,$\pm$\,0.24} & 78.14{\tiny\,$\pm$\,0.19} & 67.93{\tiny\,$\pm$\,0.24} & 76.52{\tiny\,$\pm$\,0.12} & 55.26{\tiny\,$\pm$\,0.37} & 66.75{\tiny\,$\pm$\,0.16} & 59.00{\tiny\,$\pm$\,3.17} & 73.22{\tiny\,$\pm$\,1.02} \\
    \midrule
    \emph{Person range (max--min)} & \emph{2.55{\tiny\,$\pm$\,0.91}} & \emph{0.90{\tiny\,$\pm$\,0.28}} & \emph{1.37{\tiny\,$\pm$\,0.22}} & \emph{0.80{\tiny\,$\pm$\,0.24}} & \emph{0.70{\tiny\,$\pm$\,0.15}} & \emph{0.41{\tiny\,$\pm$\,0.08}} & \emph{0.00{\tiny\,$\pm$\,0.00}} & \emph{0.00{\tiny\,$\pm$\,0.00}} \\
    \bottomrule
  \end{tabular*}
\end{table}

\subsubsection{Query dependence after domain adaptation}
\label{sec:exp:finetuned-query}

Under-canopy imagery exhibits visual characteristics that are rarely
represented in the web-scale corpora used for OVD pretraining
\citep{shao2019objects365,kamath2021mdetr,chen2023pali}: search targets
appear under severe occlusion and camouflage against cluttered
forest-floor backgrounds. This underrepresentation is consistent with the
limited zero-shot performance in Section~\ref{sec:exp:singletons} and makes domain adaptation a natural
step for improving detection performance. We thus fine-tune each OVD
with LoRA using the full-category query and re-apply the
query-composition protocol, examining not only the accuracy
gains but also whether query-composition sensitivity persists.

Comparing Tables~\ref{tab:zeroshot-forest} and~\ref{tab:zs-vs-ft}, adaptation
improves \emph{person} AP for all OVDs and yields larger gains for belongings,
mainly in AP rather than AR.
The combination of high zero-shot AR and low AP indicates
that predicted boxes were already distributed around the ground-truth
belongings but were poorly scored. The AP-dominant gains are therefore
consistent with improved alignment between the text queries and the
encoded under-canopy image features.
However, Table~\ref{tab:finetuned-ap-change} shows that it does not eliminate
query-composition sensitivity: \emph{person} AP still changes across
companion-query variants for most detectors. Thus deployment-oriented
evaluation should assess not only aggregate AP and AR, but also consistency of
individual detections across query compositions (Figure~\ref{fig:query-grid}).

\subsubsection{Instance-level query consistency}
\label{sec:exp:finetuned-turnover}
\label{sec:exp:query-consistency}

When assessing OVDs for prospective deployment in missing-person search,
aggregate detection performance should be considered together with instance-level
consistency across operationally relevant query compositions. Companion-query variants introduce additional
search targets without changing the meaning of \emph{person}; a robust OVD
should therefore preserve existing person detections as other categories are
added. NFR measures lost person detections, PFR
measures newly recovered detections, and SR measures cases in which an added
category displaces the shared target.

The instance-level results in Table~\ref{tab:query-turnover} reveal differences
that are obscured by aggregate metrics. GroundingDINO records the lowest SR
among the fusion-based detectors both zero-shot and after adaptation: its person evidence is seldom reassigned to a
belongings category. GLIP is the most query-sensitive detector, in a
favorable direction; after adaptation it combines the lowest fusion-based
NFR with the highest PFR, so companion categories
recover most persons missed by the \texttt{person}-only query while rarely
suppressing existing detections. OmDet-Turbo shows the opposite profile: it
records the highest NFR in both settings, its PFR is the only one that
decreases with adaptation, and its SR is the highest, indicating stronger confusion between \emph{person} and
belongings than in the other detectors, whereas OWLv2 remains
unchanged across all companion queries. These results suggest that SAR model selection should favor
OVDs that combine high AP and AR with low NFR and SR across query compositions.
These contrasts persist at deployment operating thresholds (Appendix~\ref{app:thresholded-consistency}).

\begin{table}[!t]
  \caption{Instance-level prediction consistency for \emph{person}
  detections under companion-query variants (ZS: zero-shot; FT:
  fine-tuned).}
  \label{tab:query-turnover}
  \centering
  \footnotesize
  \setlength{\tabcolsep}{2.0pt}
  \begin{tabular*}{\textwidth}{@{\extracolsep{\fill}}l cc cc cc cc@{}}
    \toprule
    & \multicolumn{2}{c}{GroundingDINO}
    & \multicolumn{2}{c}{GLIP}
    & \multicolumn{2}{c}{OmDet-Turbo}
    & \multicolumn{2}{c}{OWLv2} \\
    \cmidrule(lr){2-3} \cmidrule(lr){4-5}
    \cmidrule(lr){6-7} \cmidrule(lr){8-9}
    Prediction consistency metrics (\%) & ZS & FT & ZS & FT & ZS & FT & ZS & FT \\
    \midrule
    Mean NFR ($|D(q_0)|$) $\downarrow$ & 0.409 & 0.131{\tiny\,$\pm$\,0.016} & 0.363 & 0.019{\tiny\,$\pm$\,0.005} & 0.820 & 0.503{\tiny\,$\pm$\,0.065} & 0.000 & 0.000{\tiny\,$\pm$\,0.000} \\
    Mean PFR ($N-|D(q_0)|$) $\uparrow$ & 13.437 & 19.793{\tiny\,$\pm$\,1.521} & 55.388 & 78.523{\tiny\,$\pm$\,4.035} & 20.930 & 18.678{\tiny\,$\pm$\,1.037} & 0.000 & 0.000{\tiny\,$\pm$\,0.000} \\
    Mean SR ($N$) $\downarrow$ & 0.020 & 0.003{\tiny\,$\pm$\,0.002} & 0.076 & 0.006{\tiny\,$\pm$\,0.002} & 0.227 & 0.105{\tiny\,$\pm$\,0.020} & 0.000 & 0.000{\tiny\,$\pm$\,0.000} \\
    \bottomrule
  \end{tabular*}
\end{table}

\subsubsection{Sensitivity to target renaming}
\label{sec:exp:renaming}

In SAR operations, the query text supplied to an OVD may name the same search target differently: \emph{person} and \emph{hiker} can express the same search objective. We measure how well each detector still detects the original annotation class when the \emph{person} or \emph{clothes} text of the full-category query is replaced with such alternative names, as reported in Table~\ref{tab:renaming-main}.

Adaptation effects vary with the detector and the substituted name. Replacing \emph{person} with \texttt{missing person} lowers adapted GroundingDINO from $53.73$ to $39.27$ AP relative to zero-shot, whereas the other detectors improve. Across the person substitutes, only OmDet-Turbo improves nearly uniformly, while the remaining detectors are mixed. In contrast, all detectors improve consistently on the clothes substitutes, mirroring the adaptation effect in Section~\ref{sec:exp:finetuned-query}: the improved alignment between text queries and under-canopy image features transfers across the substituted names. Robustness to query wording should be evaluated alongside query composition; complete results are in Appendix~\ref{app:renaming}.

\begin{table}[H]
  \caption{AP under target renaming: the \emph{person} or \emph{clothes} text of the full-category query is replaced with an alternative name and predictions are scored against the original annotation class (ZS: zero-shot; FT: fine-tuned); full-category results are in Tables~\ref{tab:zeroshot-forest} and~\ref{tab:zs-vs-ft}.}
  \label{tab:renaming-main}
  \centering
  \footnotesize
  \setlength{\tabcolsep}{2.0pt}
  \begin{tabular*}{\textwidth}{@{\extracolsep{\fill}}l cc cc cc cc@{}}
    \toprule
    & \multicolumn{2}{c}{GroundingDINO}
    & \multicolumn{2}{c}{GLIP}
    & \multicolumn{2}{c}{OmDet-Turbo}
    & \multicolumn{2}{c}{OWLv2} \\
    \cmidrule(lr){2-3} \cmidrule(lr){4-5}
    \cmidrule(lr){6-7} \cmidrule(lr){8-9}
    Substituted name & ZS & FT & ZS & FT & ZS & FT & ZS & FT \\
    \midrule
    \multicolumn{9}{@{}l}{Substitutes for \emph{person}, \emph{person} AP} \\
    \texttt{hiker} & 58.59 & 61.96{\tiny\,$\pm$\,1.82} & 22.17 & 20.67{\tiny\,$\pm$\,6.48} & 52.60 & 52.34{\tiny\,$\pm$\,0.31} & 56.51 & 44.73{\tiny\,$\pm$\,5.56} \\
    \texttt{missing person} & 53.73 & 39.27{\tiny\,$\pm$\,3.44} & 55.98 & 64.84{\tiny\,$\pm$\,0.49} & 51.64 & 56.36{\tiny\,$\pm$\,0.43} & 55.65 & 57.63{\tiny\,$\pm$\,3.08} \\
    \texttt{victim} & 58.93 & 54.35{\tiny\,$\pm$\,5.20} & 49.81 & 46.22{\tiny\,$\pm$\,0.58} & 47.09 & 54.25{\tiny\,$\pm$\,0.16} & 54.52 & 55.37{\tiny\,$\pm$\,3.07} \\
    \midrule
    \multicolumn{9}{@{}l}{Substitutes for \emph{clothes}, \emph{clothes} AP} \\
    \texttt{clothing} & 3.81 & 68.40{\tiny\,$\pm$\,0.29} & 12.68 & 69.88{\tiny\,$\pm$\,0.27} & 31.82 & 56.34{\tiny\,$\pm$\,0.20} & 15.27 & 64.33{\tiny\,$\pm$\,1.33} \\
    \texttt{garment} & 31.34 & 66.87{\tiny\,$\pm$\,0.55} & 18.44 & 69.04{\tiny\,$\pm$\,0.18} & 33.71 & 51.18{\tiny\,$\pm$\,0.80} & 22.06 & 62.82{\tiny\,$\pm$\,2.18} \\
    \texttt{jacket} & 2.58 & 59.26{\tiny\,$\pm$\,0.86} & 4.78 & 67.13{\tiny\,$\pm$\,0.24} & 9.29 & 41.91{\tiny\,$\pm$\,2.92} & 4.52 & 45.20{\tiny\,$\pm$\,2.29} \\
    \bottomrule
  \end{tabular*}
\end{table}

\paragraph{Limitations.}
This study does not address how to mitigate query dependence or whether, in
actual SAR operations, model-assisted detections help responders
locate missing persons. Query-robust training and end-to-end field
validation remain future work.

\section{Conclusion}
\label{sec:conclusion}

We evaluated representative OVDs on the Under-Canopy
Person-and-Belongings Detection Benchmark to quantify their robustness
to query composition and target naming. Our results show that
predictions for a shared target can change when accompanying categories
or operationally equivalent names are altered, suggesting that query
choice may affect the likelihood of locating and rescuing missing
persons when OVDs are deployed for under-canopy SAR. We release
\dataset, a benchmark for detecting detached personal belongings in
forest environments and evaluating OVDs intended for SAR
deployment.

\begin{ack}
This work was supported by Institute of Information \& communications
Technology Planning \& Evaluation (IITP) grant funded by the Korea government
(MSIT) (No.\ RS-2022-II220021, Development of Core Technologies for Autonomous
Searching Drones).
\end{ack}

\bibliographystyle{plainnat}
\bibliography{refs}

\clearpage
\appendix
\section*{Appendix}
In summary, the appendix covers the following: Appendix~\ref{app:dataset}
reports the dataset splits,
per-class annotation counts, and image resolutions for \dataset.
Appendix~\ref{app:setup-details} documents the evaluation protocol,
model-native post-processing, trainable components, and adaptation cost.
Appendix~\ref{app:singletons} extends the query-composition analysis to
belongings for both zero-shot and fine-tuned detectors and complements the
person-focused results in the main text.
Appendix~\ref{app:consistency-grid} reports per-query prediction-consistency
results summarized in the main text. Appendix~\ref{app:thresholded-consistency}
evaluates prediction consistency at thresholded operating points.
Appendix~\ref{app:renaming} provides the complete target-renaming results
for zero-shot and adapted models. Appendix~\ref{app:person-absent} analyzes
\emph{person} false positives in scenes containing no person and complements
the qualitative failures in Figure~\ref{fig:belongings-confusions}. Finally,
Appendix~\ref{app:assets} documents the assets, dataset release terms, and
intended-use restrictions.

\section{\dataset Details}
\label{app:dataset}

\paragraph{Collection protocol.}
\dataset was collected to approximate viewpoints relevant to low-altitude
under-canopy MAV search. Detached belongings were placed in forest
environments and recorded with a handheld camera at heights comparable to
under-canopy MAV flight. The collection protocol follows the viewpoint design
of ForestPersons and maintains compatibility between the person and belongings
datasets. Images were organized by collection sequence for annotation and
subsequent split construction. Of the $31{,}897$ images, $27{,}663$ have a
resolution of $1920\times1080$, and $4{,}234$ have a resolution of
$1280\times720$.

\begin{table}[H]
  \caption{Dataset-level statistics for the sequence-disjoint splits of
  \dataset. Parentheses report sequence and image counts in
  \emph{easy}\,/\,\emph{medium}\,/\,\emph{hard}\,/\,background order, with background
  abbreviated as ``bg''; dedicated background sequences occur only in the
  training split and have no difficulty label.}
  \label{tab:dataset-stats}

  \centering
  \footnotesize
  \setlength{\tabcolsep}{3pt}
  \begin{tabular*}{0.95\textwidth}{@{\extracolsep{\fill}}
    >{\centering\arraybackslash}m{0.07\textwidth}
    >{\centering\arraybackslash}m{0.16\textwidth}
    >{\centering\arraybackslash}m{0.36\textwidth}
    >{\centering\arraybackslash}m{0.14\textwidth}
    >{\centering\arraybackslash}m{0.12\textwidth}}
    \toprule
    Split & \shortstack{Sequences\\{\tiny(\emph{easy}\,/\,\emph{medium}\,/\,\emph{hard}\,/\,bg)}}
      & \shortstack{Images\\{\tiny(\emph{easy}\,/\,\emph{medium}\,/\,\emph{hard}\,/\,bg)}}
      & Annotations & \shortstack{No-object\\frames} \\
    \midrule
    train & \shortstack{177\\{\scriptsize\sffamily(51\,/\,56\,/\,53\,/\,17)}}
      & \shortstack{22{,}984\\{\scriptsize\sffamily(5{,}206\,/\,7{,}168\,/\,8{,}449\,/\,2{,}161)}} & 27{,}802 & 3{,}022 \\
    val   & \shortstack{35\\{\scriptsize\sffamily(14\,/\,12\,/\,9\,/\,0)}}
      & \shortstack{4{,}486\\{\scriptsize\sffamily(1{,}114\,/\,1{,}553\,/\,1{,}819\,/\,0)}} & 6{,}225 & 203 \\
    test  & \shortstack{38\\{\scriptsize\sffamily(13\,/\,10\,/\,15\,/\,0)}}
      & \shortstack{4{,}427\\{\scriptsize\sffamily(1{,}121\,/\,1{,}448\,/\,1{,}858\,/\,0)}} & 5{,}804 & 198 \\
    \midrule
    total & \shortstack{250\\{\scriptsize\sffamily(78\,/\,78\,/\,77\,/\,17)}}
      & \shortstack{31{,}897\\{\scriptsize\sffamily(7{,}441\,/\,10{,}169\,/\,12{,}126\,/\,2{,}161)}} & 39{,}831 & 3{,}423 \\
    \bottomrule
  \end{tabular*}

\end{table}

\begin{table}[!b]
  \caption{Class-wise annotation statistics for \dataset.}
  \label{tab:dataset-class-stats}

  \centering
  \footnotesize
  \setlength{\tabcolsep}{3pt}
  \begin{tabular*}{0.95\textwidth}{@{\extracolsep{\fill}}lrrrrrrr}
    \toprule
    & clothes & hat & shoes & gloves & bag & person & etc \\
    \midrule
    train & 15{,}197 & 2{,}685 & 3{,}514 & 2{,}193 & 3{,}757 & 179 & 277 \\
    val   &  3{,}472 &   603 &   714 &   513 &   857 &  47 &  19 \\
    test  &  3{,}228 &   600 &   696 &   498 &   701 &  28 &  53 \\
    \midrule
    total & 21{,}897 & 3{,}888 & 4{,}924 & 3{,}204 & 5{,}315 & 254 & 349 \\
    Ratio & 55.0\% & 9.8\% & 12.4\% & 8.0\% & 13.3\% & 0.6\% & 0.9\% \\
    \bottomrule
  \end{tabular*}

\end{table}

\paragraph{Split construction.}
To reduce difficulty imbalance across splits, sequence-level difficulty was
estimated using zero-shot GroundingDINO
Swin-T OGC \citep{liu2024groundingdino} with the single-class query
\texttt{belongings.} Sequence-level AP$_{50}$ scores were divided into
tertiles labeled \emph{easy}, \emph{medium}, and \emph{hard}. The splits were
constructed at the sequence level (Tables~\ref{tab:dataset-stats} and~\ref{tab:dataset-class-stats}), with
sequences recorded at the same location grouped within the same split to
prevent visual overlap. The resulting sequence groups were assigned to the
train, validation, and test splits through stratified optimization that
balanced difficulty, category composition, season, location, and time of day.
We note that the difficulty-estimation configuration differs from the
GroundingDINO configuration evaluated in the main experiments, which uses a
Swin-B backbone with category-specific queries. The difficulty scores served
only as a balancing heuristic for split assignment, and all detectors were
trained and evaluated on the same fixed splits; the resulting labels are not
intended as model-independent difficulty annotations. 17 dedicated
background sequences containing no benchmark targets were intentionally
collected as negative training samples and were not assigned a difficulty
level. Among difficulty-scored images,
the easy, medium, and hard shares were 25.0/34.4/40.6\% for train,
24.8/34.6/40.5\% for validation, and 25.3/32.7/42.0\% for test.

\paragraph{Annotation protocol.}
Categories were assigned using records of the belongings placed during each
collection sequence together with the temporal context provided by adjacent
video frames. Bounding boxes and category labels were stored in standard COCO
format. The \texttt{iscrowd} field marks annotated boxes that are retained in
the release but ignored during training and excluded from evaluation. In total,
$696$ boxes carry this flag: all \emph{etc} annotations, which cover objects
outside the benchmark category definitions, and benchmark-category boxes for
which the square root of the box area is below $16$ pixels after the standard
detector resize (\texttt{ResizeShortestEdge}(800, max 1333)). The latter
criterion corresponds to an original-image square-root area below approximately
$23.0$ pixels at $1920\times1080$ resolution and $15.4$ pixels at
$1280\times720$ resolution.

\section{Training and Post-processing Details}
\label{app:setup-details}

\paragraph{Evaluation protocol.}

All reported AP, AR, and consistency results for the OVDs use the same
\texttt{EvalProtocol}: \texttt{score\_threshold}=0.0,
\texttt{max\_detections\_per\_image}=100,
\texttt{nms}=\texttt{model\_default}, \texttt{box\_threshold}=0.0, and
\texttt{text\_threshold}=0.0, with deterministic execution. AP and AR are
computed over the standard COCO IoU sweep from 0.50 to 0.95 in increments
of 0.05. Every evaluation entry point calls
\texttt{assert\_frozen\_protocol} before writing a result table.
Table~\ref{tab:postprocessing} summarizes the model-native post-processing,
which remains unchanged within the fixed protocol. No detector merges boxes
across queried categories or collapses the query axis with a forced
single-label argmax. Consequently, \emph{person}-to-belongings reassignments
in the flip analysis are not artifacts of decoding.

\begin{table}[H]
  \caption{Model-native post-processing retained during evaluation.}
  \label{tab:postprocessing}

  \centering
  \footnotesize
  \setlength{\tabcolsep}{3pt}
  \begin{tabular*}{0.95\textwidth}{@{\extracolsep{\fill}}l l >{\raggedright\arraybackslash}p{0.56\linewidth}@{}}
    \toprule
    OVD & Native NMS & Details \\
    \midrule
    GroundingDINO & none
      & Detectron2-style top-$k$ selection over flattened query--class scores;
        \texttt{select\_box\_nums\_for\_evaluation}=200. \\
    GLIP & batched NMS, IoU 0.6
      & \texttt{nms\_pre}=1000 with MMDetection batched NMS. The default test
        config uses \texttt{score\_thr}=0.05 and \texttt{max\_per\_img}=100;
        the evaluation overrides only these two fields to \texttt{score\_thr}=0.0 and
        \texttt{max\_per\_img}=500 so that every evaluation category can retain
        candidates before COCO scoring. \\
    OmDet-Turbo & batched NMS, IoU 0.5
      & Processor-signature default NMS with \texttt{num\_queries}=900. \\
    OWLv2 & none
      & Top-$k$ selection over flattened patch--query scores. The default
        top-$k$ is 100; the evaluation overrides
        \texttt{max\_output\_detections} to 500 so that every evaluation
        category can retain candidates before COCO scoring. \\
    \bottomrule
  \end{tabular*}

\end{table}

\paragraph{Compute.}

All adaptation runs were performed on a single NVIDIA RTX 6000 Ada GPU with
world size one. One epoch comprises $90{,}670$ training samples drawn from the
combined ForestPersons and \dataset training splits. With an effective batch
size of $16$, one epoch yields $5{,}667$ optimizer steps; the final step uses a
partial batch. Per-device batch size and gradient accumulation are adjusted by
model to maintain the same effective batch size. Table~\ref{tab:compute}
reports the resulting computational cost. Each OVD is adapted with three
random seeds to assess the statistical reliability of the fine-tuned results.
Across the three seeds, adapting all four detectors requires approximately
$60$ GPU-hours.

\begin{table}[H]
  \caption{Adaptation cost per model on one NVIDIA RTX 6000 Ada.
  \emph{Batch} gives the per-device batch size and gradient-accumulation
  factor for an effective batch size of $16$; ranges span three seeds.}
  \label{tab:compute}

  \centering
  \footnotesize
  \setlength{\tabcolsep}{4pt}
  \begin{tabular*}{0.95\textwidth}{@{\extracolsep{\fill}}l cccc@{}}
    \toprule
    & GroundingDINO & GLIP & OmDet-Turbo & OWLv2 \\
    & Swin-B & T(C) & Swin-T & large \\
    \midrule
    Params (M) & 235.7 & 234.6 & 116.2 & 444.7 \\
    Batch (batch $\times$ accum.) & $2\times8$ & $2\times8$ & $8\times2$ & $4\times4$ \\
    Time (s/step) & 2.815--2.971 & 4.439--4.519 & 0.245--0.246 & 5.084--5.141 \\
    Wall-clock, one run (h) & 4.43--4.68 & 6.99--7.11 & 0.39 & 8.00--8.09 \\
    Wall-clock, three seeds (h) & 13.7 & 21.2 & 1.2 & 24.1 \\
    \bottomrule
  \end{tabular*}

\end{table}

\paragraph{Trainable components.}

In the following component descriptions, \emph{adapting} a component means inserting
LoRA adapters into the component, whereas detection heads described as trained are
updated in full. For GroundingDINO, LoRA adapters are inserted only into the
72 feature-enhancer modules, while the remaining modules, including the
pretrained text encoder, remain frozen. For GLIP, LoRA adapters are inserted
into the DyHead tower, including the multimodal fusion and in-tower text
pathways, without updating the base text encoder. For OmDet-Turbo, only the
decoder is adapted, together with the trainable box and classification heads.
The OmDet-Turbo encoder remains frozen, preserving the pretrained pathway that
forms class-query representations. For OWLv2, LoRA adapters are inserted into
the vision encoder, and the box and class heads are trained in full; the text
encoder and objectness head remain frozen. Overall, LoRA updates are
concentrated in the visual and downstream cross-modal detection pathways,
while the base text encoders remain frozen. The adaptation design preserves
the pretrained lexical representations and focuses domain learning on the
visual characteristics of belongings in under-canopy imagery. Because several
detectors adapt cross-modal fusion modules, the design does not guarantee
invariant cross-modal interactions across query compositions.

\section{Single-Category and Companion-Query Results}
\label{app:singletons}

Table~\ref{tab:singletons-app} expands the per-category belongings results
deferred from Section~\ref{sec:exp:singletons}. Each belongings category is
evaluated as a singleton query, paired with \emph{person}, and included in the
full-category query. The three conditions mirror the query-composition
protocol applied to the \texttt{person} reference query in the main text.
The paired queries correspond to Table~\ref{tab:pair-identity}, and the
full-category rows reproduce Table~\ref{tab:zeroshot-forest}.

The belongings results exhibit the same query-composition dependence as the
person results. For the fusion-based detectors, AP can differ by tens of points
between scoring a class alone and scoring the same class within a larger query.
The direction of the difference depends on the class and detector:
GroundingDINO loses \emph{shoes} AP as the query vocabulary grows, whereas
OmDet-Turbo gains \emph{clothes} AP under the full-category query. OWLv2 is
nearly unchanged across the three conditions, consistent with largely
independent per-term scoring.

\begin{table}[H]
  \caption{Class-wise belongings AP and AR for \textbf{pretrained zero-shot}
  OVDs when each class is queried alone, paired with \emph{person}, or included
  in the full-category query.}
  \label{tab:singletons-app}

  \centering
  \footnotesize
  \setlength{\tabcolsep}{1.6pt}
  \begin{tabular*}{0.95\textwidth}{@{\extracolsep{\fill}}ll cc cc cc cc@{}}
    \toprule
    & & \multicolumn{2}{c}{GroundingDINO} & \multicolumn{2}{c}{GLIP}
      & \multicolumn{2}{c}{OmDet-Turbo} & \multicolumn{2}{c}{OWLv2} \\
    \cmidrule(lr){3-4} \cmidrule(lr){5-6} \cmidrule(lr){7-8} \cmidrule(lr){9-10}
    Class & Text query & AP & AR & AP & AR & AP & AR & AP & AR \\
    \midrule
    \multirow{3}{*}{\texttt{clothes}} & \texttt{clothes} & 10.84 & 77.20 & \phantom{0}9.56 & 52.14 & 22.97 & 64.64 & 24.37 & 80.22 \\
       & \texttt{person clothes} & \phantom{0}7.09 & 77.29 & 11.40 & 72.30 & 20.13 & 67.09 & 24.37 & 79.75 \\
       & \emph{full-category query} & \phantom{0}4.12 & 78.90 & 11.97 & 71.72 & 33.61 & 72.55 & 24.37 & 79.00 \\
    \midrule
    \multirow{3}{*}{\texttt{hat}} & \texttt{hat} & \phantom{0}9.30 & 85.80 & \phantom{0}8.04 & 57.87 & \phantom{0}1.69 & 61.98 & \phantom{0}7.17 & 82.65 \\
       & \texttt{person hat} & \phantom{0}1.65 & 85.07 & \phantom{0}2.10 & 80.07 & \phantom{0}2.34 & 64.28 & \phantom{0}7.17 & 82.42 \\
       & \emph{full-category query} & \phantom{0}0.84 & 85.40 & \phantom{0}1.73 & 79.42 & \phantom{0}1.06 & 72.92 & \phantom{0}7.18 & 82.18 \\
    \midrule
    \multirow{3}{*}{\texttt{shoes}} & \texttt{shoes} & 22.79 & 80.91 & 11.00 & 48.23 & \phantom{0}7.12 & 72.99 & 24.22 & 82.08 \\
       & \texttt{person shoes} & 13.12 & 79.81 & \phantom{0}5.95 & 71.21 & \phantom{0}7.25 & 72.26 & 24.22 & 82.08 \\
       & \emph{full-category query} & \phantom{0}6.78 & 78.59 & \phantom{0}1.96 & 66.85 & \phantom{0}9.57 & 71.29 & 24.22 & 81.97 \\
    \midrule
    \multirow{3}{*}{\texttt{gloves}} & \texttt{gloves} & \phantom{0}4.73 & 76.97 & \phantom{0}4.85 & 57.65 & \phantom{0}3.74 & 69.22 & \phantom{0}6.54 & 78.37 \\
       & \texttt{person gloves} & \phantom{0}1.67 & 76.02 & \phantom{0}3.30 & 71.55 & \phantom{0}2.64 & 69.08 & \phantom{0}6.54 & 78.29 \\
       & \emph{full-category query} & \phantom{0}0.55 & 76.53 & \phantom{0}1.59 & 67.53 & \phantom{0}1.03 & 69.14 & \phantom{0}6.54 & 77.95 \\
    \midrule
    \multirow{3}{*}{\texttt{bag}} & \texttt{bag} & 17.99 & 77.75 & \phantom{0}7.90 & 53.85 & 10.74 & 67.43 & 25.80 & 78.15 \\
       & \texttt{person bag} & 16.06 & 76.09 & 10.68 & 73.65 & 10.91 & 65.49 & 25.80 & 78.00 \\
       & \emph{full-category query} & 20.06 & 74.54 & \phantom{0}5.12 & 72.34 & 12.35 & 68.25 & 25.80 & 77.75 \\
    \midrule
    \multirow{3}{*}{\shortstack[l]{\emph{Belongings}\\\emph{mean}}} & \emph{each class alone} & \emph{13.13} & \emph{79.72} & \emph{\phantom{0}8.27} & \emph{53.95} & \emph{\phantom{0}9.25} & \emph{67.25} & \emph{17.62} & \emph{80.29} \\
       & \emph{each class with} \texttt{person} & \emph{\phantom{0}7.92} & \emph{78.86} & \emph{\phantom{0}6.69} & \emph{73.75} & \emph{\phantom{0}8.65} & \emph{67.64} & \emph{17.62} & \emph{80.11} \\
       & \emph{full-category query} & \emph{\phantom{0}6.47} & \emph{78.79} & \emph{\phantom{0}4.47} & \emph{71.57} & \emph{11.52} & \emph{70.83} & \emph{17.62} & \emph{79.77} \\
    \bottomrule
  \end{tabular*}

\end{table}

Table~\ref{tab:singletons-app-ft} repeats the three query conditions for the LoRA fine-tuned detectors. Adaptation raises belongings AP in every condition, yet for the fusion-based detectors the query dependence does not disappear; it becomes directional. AP now increases, almost without exception, from the singleton query to the \emph{person} pair to the full-category query presented during adaptation; for GLIP, \emph{clothes} rises from $22.13$ AP under the singleton query to $70.26$ inside the full-category query. Because adaptation always paired training images with the full-category query, this ordering indicates that fine-tuning aligns belongings scoring with the trained query composition rather than making it query-invariant. OWLv2 again remains essentially unchanged across the three conditions.

\begin{table}[!tp]
  \caption{Class-wise belongings AP and AR for \textbf{LoRA fine-tuned} OVDs under the same three query conditions as Table~\ref{tab:singletons-app}: each class scored alone, paired with \emph{person}, and inside the full-category query used during adaptation. The full-category rows reproduce Table~\ref{tab:zs-vs-ft}; entries are the mean over three adaptation seeds.}
  \label{tab:singletons-app-ft}

  \centering
  \footnotesize
  \setlength{\tabcolsep}{0.6pt}
  \begin{tabular*}{\textwidth}{@{\extracolsep{\fill}}ll cc cc cc cc@{}}
    \toprule
    & & \multicolumn{2}{c}{GroundingDINO} & \multicolumn{2}{c}{GLIP}
      & \multicolumn{2}{c}{OmDet-Turbo} & \multicolumn{2}{c}{OWLv2} \\
    \cmidrule(lr){3-4} \cmidrule(lr){5-6} \cmidrule(lr){7-8} \cmidrule(lr){9-10}
    Class & Text query & AP & AR & AP & AR & AP & AR & AP & AR \\
    \midrule
       \multirow{3}{*}{\texttt{clothes}} & \texttt{clothes} & 54.41{\tiny$\pm$0.67} & 82.91{\tiny$\pm$0.85} & 22.13{\tiny$\pm$1.84} & 68.09{\tiny$\pm$2.23} & 51.15{\tiny$\pm$0.95} & 72.96{\tiny$\pm$0.32} & 64.74{\tiny$\pm$1.35} & 79.48{\tiny$\pm$0.79} \\
        & \texttt{person clothes} & 60.84{\tiny$\pm$0.17} & 83.04{\tiny$\pm$0.34} & 60.13{\tiny$\pm$0.42} & 82.87{\tiny$\pm$0.24} & 51.10{\tiny$\pm$0.62} & 72.46{\tiny$\pm$0.26} & 64.74{\tiny$\pm$1.35} & 79.48{\tiny$\pm$0.79} \\
        & \emph{full-category query} & 68.97{\tiny$\pm$0.29} & 83.67{\tiny$\pm$0.13} & 70.26{\tiny$\pm$0.14} & 83.21{\tiny$\pm$0.06} & 56.02{\tiny$\pm$0.28} & 74.70{\tiny$\pm$0.15} & 64.74{\tiny$\pm$1.35} & 79.45{\tiny$\pm$0.77} \\
    \midrule
       \multirow{3}{*}{\texttt{hat}} & \texttt{hat} & 19.12{\tiny$\pm$1.11} & 87.54{\tiny$\pm$0.32} & 19.78{\tiny$\pm$2.49} & 74.96{\tiny$\pm$1.28} & 24.98{\tiny$\pm$2.08} & 71.44{\tiny$\pm$1.19} & 58.28{\tiny$\pm$1.91} & 82.47{\tiny$\pm$0.62} \\
        & \texttt{person hat} & 28.66{\tiny$\pm$0.94} & 87.13{\tiny$\pm$0.44} & 38.73{\tiny$\pm$3.50} & 82.21{\tiny$\pm$0.44} & 24.24{\tiny$\pm$1.66} & 71.09{\tiny$\pm$1.00} & 58.28{\tiny$\pm$1.91} & 82.47{\tiny$\pm$0.62} \\
        & \emph{full-category query} & 53.26{\tiny$\pm$0.76} & 85.72{\tiny$\pm$0.68} & 56.49{\tiny$\pm$1.25} & 84.09{\tiny$\pm$0.61} & 37.33{\tiny$\pm$1.31} & 74.99{\tiny$\pm$0.39} & 58.28{\tiny$\pm$1.91} & 82.41{\tiny$\pm$0.55} \\
    \midrule
       \multirow{3}{*}{\texttt{shoes}} & \texttt{shoes} & 32.32{\tiny$\pm$1.46} & 82.61{\tiny$\pm$0.62} & 13.51{\tiny$\pm$2.06} & 61.96{\tiny$\pm$1.94} & 31.44{\tiny$\pm$1.04} & 74.54{\tiny$\pm$0.58} & 57.84{\tiny$\pm$1.86} & 80.65{\tiny$\pm$0.52} \\
        & \texttt{person shoes} & 42.01{\tiny$\pm$1.09} & 82.61{\tiny$\pm$0.76} & 39.63{\tiny$\pm$1.85} & 83.05{\tiny$\pm$0.69} & 34.23{\tiny$\pm$1.25} & 74.07{\tiny$\pm$0.39} & 57.84{\tiny$\pm$1.86} & 80.65{\tiny$\pm$0.52} \\
        & \emph{full-category query} & 54.25{\tiny$\pm$0.08} & 84.20{\tiny$\pm$0.07} & 54.38{\tiny$\pm$1.80} & 83.87{\tiny$\pm$0.54} & 35.53{\tiny$\pm$1.44} & 73.66{\tiny$\pm$0.41} & 57.84{\tiny$\pm$1.86} & 80.63{\tiny$\pm$0.52} \\
    \midrule
       \multirow{3}{*}{\texttt{gloves}} & \texttt{gloves} & 13.78{\tiny$\pm$0.99} & 77.40{\tiny$\pm$0.66} & 10.76{\tiny$\pm$2.62} & 65.82{\tiny$\pm$3.20} & 17.58{\tiny$\pm$1.73} & 73.36{\tiny$\pm$0.94} & 37.87{\tiny$\pm$1.52} & 77.70{\tiny$\pm$1.98} \\
        & \texttt{person gloves} & 26.01{\tiny$\pm$2.19} & 76.21{\tiny$\pm$0.32} & 29.22{\tiny$\pm$3.53} & 79.40{\tiny$\pm$0.85} & 19.23{\tiny$\pm$2.52} & 72.35{\tiny$\pm$0.30} & 37.87{\tiny$\pm$1.52} & 77.69{\tiny$\pm$1.97} \\
        & \emph{full-category query} & 34.29{\tiny$\pm$1.36} & 79.33{\tiny$\pm$0.64} & 29.72{\tiny$\pm$3.20} & 78.50{\tiny$\pm$0.64} & 21.89{\tiny$\pm$1.45} & 72.10{\tiny$\pm$0.36} & 37.87{\tiny$\pm$1.52} & 77.42{\tiny$\pm$1.84} \\
    \midrule
       \multirow{3}{*}{\texttt{bag}} & \texttt{bag} & 25.73{\tiny$\pm$0.93} & 79.71{\tiny$\pm$0.65} & 13.39{\tiny$\pm$1.30} & 58.26{\tiny$\pm$1.55} & 31.26{\tiny$\pm$0.69} & 68.03{\tiny$\pm$0.34} & 54.27{\tiny$\pm$0.32} & 76.42{\tiny$\pm$0.65} \\
        & \texttt{person bag} & 32.09{\tiny$\pm$2.34} & 79.86{\tiny$\pm$0.44} & 25.94{\tiny$\pm$1.31} & 73.40{\tiny$\pm$0.49} & 31.50{\tiny$\pm$0.60} & 67.27{\tiny$\pm$0.15} & 54.27{\tiny$\pm$0.32} & 76.42{\tiny$\pm$0.65} \\
        & \emph{full-category query} & 48.84{\tiny$\pm$1.80} & 79.71{\tiny$\pm$0.22} & 35.76{\tiny$\pm$1.85} & 76.09{\tiny$\pm$0.83} & 33.51{\tiny$\pm$0.45} & 68.02{\tiny$\pm$0.24} & 54.27{\tiny$\pm$0.32} & 76.40{\tiny$\pm$0.69} \\
    \midrule
       \multirow{3}{*}{\shortstack[l]{\emph{Belongings}\\\emph{mean}}} & \emph{each class alone} & \emph{29.07{\tiny$\pm$0.40}} & \emph{82.03{\tiny$\pm$0.33}} & \emph{15.91{\tiny$\pm$0.99}} & \emph{65.82{\tiny$\pm$1.64}} & \emph{31.28{\tiny$\pm$0.20}} & \emph{72.06{\tiny$\pm$0.46}} & \emph{54.60{\tiny$\pm$1.24}} & \emph{79.34{\tiny$\pm$0.53}} \\
        & \emph{each class with} \texttt{person} & \emph{37.92{\tiny$\pm$0.55}} & \emph{81.77{\tiny$\pm$0.27}} & \emph{38.73{\tiny$\pm$0.33}} & \emph{80.19{\tiny$\pm$0.31}} & \emph{32.06{\tiny$\pm$0.62}} & \emph{71.45{\tiny$\pm$0.25}} & \emph{54.60{\tiny$\pm$1.24}} & \emph{79.34{\tiny$\pm$0.53}} \\
        & \emph{full-category query} & \emph{51.92{\tiny$\pm$0.45}} & \emph{82.53{\tiny$\pm$0.02}} & \emph{49.32{\tiny$\pm$0.46}} & \emph{81.15{\tiny$\pm$0.35}} & \emph{36.85{\tiny$\pm$0.40}} & \emph{72.70{\tiny$\pm$0.04}} & \emph{54.60{\tiny$\pm$1.24}} & \emph{79.26{\tiny$\pm$0.50}} \\
    \bottomrule
  \end{tabular*}

\end{table}

\FloatBarrier

\section{Per-Query Prediction Consistency}
\label{app:consistency-grid}

\suppressfloats[t]

To examine how individual query compositions affect \emph{person}
predictions, Table~\ref{tab:query-turnover-full} reports NFR, PFR, and SR for
each companion-query variant. The flip rates are reported in both conditional
and benchmark-normalized forms. Here, $N=20{,}895$ is the benchmark-wide
ground-truth person count, and $|D(q_0)|$ is the number of ground-truth person
instances detected by $q_0$ at IoU $\geq 0.5$. Normalizing NFR by $|D(q_0)|$ measures the
fraction of person instances detected by the reference query that are lost
under a companion query. Normalizing PFR by $N-|D(q_0)|$ measures the fraction
of previously missed person instances that are recovered. Normalizing either
rate by $N$ places all companion-query variants on a common benchmark-wide
scale. The model contrasts hold for every individual query variant rather than
being driven by a single variant. OmDet-Turbo records the highest NFR and SR
for every companion query, whereas OWLv2 is invariant across all queries and
metrics. For zero-shot GroundingDINO, the
\texttt{person clothes} query provides the highest recovery among the
companion-query variants, reaching $23.4\%$ of the missed persons compared with
at most $16.0\%$ for the other variants. After adaptation, recovery improves
for most GroundingDINO variants and peaks under the full-category query at
$26.8\%$. GLIP exhibits the strongest adaptation effect: recovery is uniformly
high at about $79\%$ across all variants, while NFR decreases by an order of
magnitude from $0.28$--$0.47\%$ to $0.01$--$0.02\%$. The GLIP results indicate
that companion queries recover additional person detections while rarely
suppressing existing detections. In contrast, OmDet-Turbo shows slightly lower
recovery after adaptation across the companion-query variants.

\begin{table}[!tp]
  \caption{Per-query prediction-consistency metrics for \emph{person}
  relative to $q_0=\texttt{person}$ (ZS: zero-shot; FT: fine-tuned). Values are
  percentages; FT entries report the mean and sample standard deviation over
  three adaptation seeds.}
  \label{tab:query-turnover-full}

  \centering
  \footnotesize
  \setlength{\tabcolsep}{1.1pt}
  \begin{tabular*}{0.97\textwidth}{@{\extracolsep{\fill}}l cc cc cc cc@{}}
    \toprule
    & \multicolumn{2}{c}{GroundingDINO}
    & \multicolumn{2}{c}{GLIP}
    & \multicolumn{2}{c}{OmDet-Turbo}
    & \multicolumn{2}{c}{OWLv2} \\
    \cmidrule(lr){2-3} \cmidrule(lr){4-5}
    \cmidrule(lr){6-7} \cmidrule(lr){8-9}
    Text query & ZS & FT & ZS & FT & ZS & FT & ZS & FT \\
    \midrule
    \multicolumn{9}{@{}l}{\emph{Negative flip rate} (NFR), normalized by $|D(q_0)|$~$\downarrow$} \\
    \texttt{person clothes} & 0.293 & 0.196{\tiny$\pm$0.018} & 0.352 & 0.016{\tiny$\pm$0.003} & 0.729 & 0.625{\tiny$\pm$0.051} & 0.000 & 0.000{\tiny$\pm$0.000} \\
    \texttt{person hat} & 0.381 & 0.143{\tiny$\pm$0.022} & 0.308 & 0.021{\tiny$\pm$0.006} & 0.705 & 0.363{\tiny$\pm$0.037} & 0.000 & 0.000{\tiny$\pm$0.000} \\
    \texttt{person shoes} & 0.278 & 0.103{\tiny$\pm$0.040} & 0.348 & 0.020{\tiny$\pm$0.008} & 1.020 & 0.401{\tiny$\pm$0.086} & 0.000 & 0.000{\tiny$\pm$0.000} \\
    \texttt{person gloves} & 0.396 & 0.092{\tiny$\pm$0.013} & 0.279 & 0.013{\tiny$\pm$0.006} & 0.660 & 0.338{\tiny$\pm$0.071} & 0.000 & 0.000{\tiny$\pm$0.000} \\
    \texttt{person bag} & 0.210 & 0.109{\tiny$\pm$0.016} & 0.426 & 0.023{\tiny$\pm$0.006} & 0.729 & 0.542{\tiny$\pm$0.074} & 0.000 & 0.000{\tiny$\pm$0.000} \\
    \emph{full-category query} & 0.899 & 0.143{\tiny$\pm$0.003} & 0.465 & 0.021{\tiny$\pm$0.006} & 1.074 & 0.750{\tiny$\pm$0.081} & 0.000 & 0.000{\tiny$\pm$0.000} \\
    \midrule
    \multicolumn{9}{@{}l}{\emph{Negative flip rate} (NFR), normalized by $N$~$\downarrow$} \\
    \texttt{person clothes} & 0.287 & 0.195{\tiny$\pm$0.018} & 0.345 & 0.016{\tiny$\pm$0.003} & 0.708 & 0.608{\tiny$\pm$0.050} & 0.000 & 0.000{\tiny$\pm$0.000} \\
    \texttt{person hat} & 0.373 & 0.142{\tiny$\pm$0.022} & 0.302 & 0.021{\tiny$\pm$0.006} & 0.684 & 0.353{\tiny$\pm$0.036} & 0.000 & 0.000{\tiny$\pm$0.000} \\
    \texttt{person shoes} & 0.273 & 0.102{\tiny$\pm$0.040} & 0.340 & 0.019{\tiny$\pm$0.008} & 0.991 & 0.389{\tiny$\pm$0.084} & 0.000 & 0.000{\tiny$\pm$0.000} \\
    \texttt{person gloves} & 0.388 & 0.091{\tiny$\pm$0.013} & 0.273 & 0.013{\tiny$\pm$0.006} & 0.641 & 0.329{\tiny$\pm$0.070} & 0.000 & 0.000{\tiny$\pm$0.000} \\
    \texttt{person bag} & 0.206 & 0.108{\tiny$\pm$0.015} & 0.416 & 0.022{\tiny$\pm$0.006} & 0.708 & 0.526{\tiny$\pm$0.072} & 0.000 & 0.000{\tiny$\pm$0.000} \\
    \emph{full-category query} & 0.881 & 0.142{\tiny$\pm$0.003} & 0.455 & 0.021{\tiny$\pm$0.006} & 1.043 & 0.729{\tiny$\pm$0.080} & 0.000 & 0.000{\tiny$\pm$0.000} \\
    \midrule
    \multicolumn{9}{@{}l}{\emph{Positive flip rate} (PFR), normalized by $N-|D(q_0)|$~$\uparrow$} \\
    \texttt{person clothes} & 23.445 & 22.957{\tiny$\pm$1.172} & 55.388 & 78.814{\tiny$\pm$3.687} & 22.425 & 20.775{\tiny$\pm$1.285} & 0.000 & 0.000{\tiny$\pm$0.000} \\
    \texttt{person hat} & 11.005 & 20.787{\tiny$\pm$2.604} & 57.112 & 79.093{\tiny$\pm$4.290} & 18.439 & 16.645{\tiny$\pm$1.123} & 0.000 & 0.000{\tiny$\pm$0.000} \\
    \texttt{person shoes} & 7.656 & 16.145{\tiny$\pm$1.611} & 55.172 & 78.536{\tiny$\pm$3.755} & 21.262 & 16.688{\tiny$\pm$0.520} & 0.000 & 0.000{\tiny$\pm$0.000} \\
    \texttt{person gloves} & 12.201 & 16.669{\tiny$\pm$1.903} & 56.897 & 78.244{\tiny$\pm$3.710} & 18.605 & 15.967{\tiny$\pm$1.711} & 0.000 & 0.000{\tiny$\pm$0.000} \\
    \texttt{person bag} & 10.287 & 15.437{\tiny$\pm$2.001} & 56.466 & 78.171{\tiny$\pm$4.597} & 20.930 & 18.399{\tiny$\pm$2.005} & 0.000 & 0.000{\tiny$\pm$0.000} \\
    \emph{full-category query} & 16.029 & 26.765{\tiny$\pm$5.041} & 51.293 & 78.280{\tiny$\pm$4.360} & 23.920 & 23.592{\tiny$\pm$0.867} & 0.000 & 0.000{\tiny$\pm$0.000} \\
    \midrule
    \multicolumn{9}{@{}l}{\emph{Positive flip rate} (PFR), normalized by $N$~$\uparrow$} \\
    \texttt{person clothes} & 0.469 & 0.204{\tiny$\pm$0.006} & 1.230 & 1.777{\tiny$\pm$0.514} & 0.646 & 0.585{\tiny$\pm$0.018} & 0.000 & 0.000{\tiny$\pm$0.000} \\
    \texttt{person hat} & 0.220 & 0.185{\tiny$\pm$0.024} & 1.268 & 1.785{\tiny$\pm$0.530} & 0.531 & 0.469{\tiny$\pm$0.017} & 0.000 & 0.000{\tiny$\pm$0.000} \\
    \texttt{person shoes} & 0.153 & 0.144{\tiny$\pm$0.013} & 1.225 & 1.771{\tiny$\pm$0.513} & 0.613 & 0.471{\tiny$\pm$0.003} & 0.000 & 0.000{\tiny$\pm$0.000} \\
    \texttt{person gloves} & 0.244 & 0.148{\tiny$\pm$0.017} & 1.263 & 1.764{\tiny$\pm$0.512} & 0.536 & 0.450{\tiny$\pm$0.039} & 0.000 & 0.000{\tiny$\pm$0.000} \\
    \texttt{person bag} & 0.206 & 0.137{\tiny$\pm$0.015} & 1.254 & 1.766{\tiny$\pm$0.533} & 0.603 & 0.518{\tiny$\pm$0.048} & 0.000 & 0.000{\tiny$\pm$0.000} \\
    \emph{full-category query} & 0.321 & 0.238{\tiny$\pm$0.041} & 1.139 & 1.768{\tiny$\pm$0.528} & 0.689 & 0.665{\tiny$\pm$0.008} & 0.000 & 0.000{\tiny$\pm$0.000} \\
    \midrule
    \multicolumn{9}{@{}l}{\emph{Stealing rate} (SR), normalized by $N$~$\downarrow$} \\
    \texttt{person clothes} & 0.019 & 0.005{\tiny$\pm$0.005} & 0.038 & 0.005{\tiny$\pm$0.000} & 0.244 & 0.140{\tiny$\pm$0.035} & 0.000 & 0.000{\tiny$\pm$0.000} \\
    \texttt{person hat} & 0.000 & 0.003{\tiny$\pm$0.003} & 0.062 & 0.006{\tiny$\pm$0.003} & 0.177 & 0.075{\tiny$\pm$0.030} & 0.000 & 0.000{\tiny$\pm$0.000} \\
    \texttt{person shoes} & 0.005 & 0.002{\tiny$\pm$0.003} & 0.024 & 0.006{\tiny$\pm$0.003} & 0.105 & 0.062{\tiny$\pm$0.014} & 0.000 & 0.000{\tiny$\pm$0.000} \\
    \texttt{person gloves} & 0.005 & 0.000{\tiny$\pm$0.000} & 0.019 & 0.005{\tiny$\pm$0.000} & 0.206 & 0.086{\tiny$\pm$0.030} & 0.000 & 0.000{\tiny$\pm$0.000} \\
    \texttt{person bag} & 0.000 & 0.002{\tiny$\pm$0.003} & 0.168 & 0.005{\tiny$\pm$0.005} & 0.235 & 0.085{\tiny$\pm$0.019} & 0.000 & 0.000{\tiny$\pm$0.000} \\
    \emph{full-category query} & 0.091 & 0.005{\tiny$\pm$0.005} & 0.144 & 0.008{\tiny$\pm$0.003} & 0.392 & 0.180{\tiny$\pm$0.035} & 0.000 & 0.000{\tiny$\pm$0.000} \\
    \bottomrule
  \end{tabular*}

\end{table}

\FloatBarrier

\section{Thresholded Prediction Consistency at Operating Points}
\label{app:thresholded-consistency}

The consistency metrics above score all predictions. A negative flip at score
threshold $0.0$ therefore indicates that a person instance disappears from the
ranked output entirely. Under the assumed deployment setting, detections act as
alerts only above an operating threshold. A person instance whose score falls
below the operating threshold under a companion query would no longer trigger
an alert, even if the instance remains in the ranked output. The thresholded
analysis connects the ranking-level and alert-level views. An instance counts
as detected under a query only if a same-class prediction with score
$\geq \tau$ matches the instance at IoU $\geq 0.5$, and $D_\tau(q_0)$ replaces
$D(q_0)$ in the denominators. Setting $\tau=0$ recovers the main protocol,
while $\tau \in \{0.1, 0.25, 0.5\}$ evaluates consistency at increasingly
selective operating points. Table~\ref{tab:query-turnover-thresholded}
reports the resulting mean and maximum over the six companion-query variants,
computed within each seed and then averaged. The symbol $^{\ddag}$ marks an
undefined rate when the reference query yields no detections above the threshold.

Raising the threshold converts ranking-level changes into operational losses.
Across the reported operating thresholds, mean NFR$_\tau$ increases with
$\tau$ for every fusion-based detector. Fine-tuned mean NFR$_\tau$
nevertheless remains below $1.7\%$ even at $\tau=0.5$. OmDet-Turbo records the
highest mean NFR$_\tau$ at every threshold. GLIP combines the lowest
fusion-based mean NFR$_\tau$ with by far the highest mean PFR$_\tau$, whereas
OWLv2 remains at or below $0.002\%$ when NFR$_\tau$ is defined. Adaptation
substantially reduces high-threshold losses. At $\tau=0.5$, zero-shot
OmDet-Turbo loses up to $36\%$ of the reference detections, whereas the adapted
model loses at most $2.3\%$. In addition, SR$_\tau$ can exceed the
unthresholded value because thresholding removes person detections whose
locations remain covered by higher-scoring belongings predictions. The
increase is consistent with the class-assignment interpretation in
Appendix~\ref{app:person-absent}. Together, the thresholded results indicate
that the query-composition sensitivity reported in the main text is not an
artifact of the unthresholded protocol. Losses become larger at deployment
operating points, reinforcing the need for consistency-aware model selection.

\begin{table}[!tp]
  \caption{Thresholded prediction-consistency metrics for \emph{person}
  relative to $q_0=\texttt{person}$ (ZS: zero-shot; FT: fine-tuned). Mean and
  Max are computed over six companion-query variants; FT entries summarize
  three adaptation seeds. $^{\ddag}$ indicates an undefined rate.}
  \label{tab:query-turnover-thresholded}

  \centering
  \footnotesize
  \setlength{\tabcolsep}{2.0pt}
  \begin{tabular*}{0.97\textwidth}{@{\extracolsep{\fill}}l cc cc cc cc@{}}
    \toprule
    & \multicolumn{2}{c}{GroundingDINO}
    & \multicolumn{2}{c}{GLIP}
    & \multicolumn{2}{c}{OmDet-Turbo}
    & \multicolumn{2}{c}{OWLv2} \\
    \cmidrule(lr){2-3} \cmidrule(lr){4-5}
    \cmidrule(lr){6-7} \cmidrule(lr){8-9}
    Prediction consistency metrics (\%) & ZS & FT & ZS & FT & ZS & FT & ZS & FT \\
    \midrule
    \multicolumn{9}{@{}l}{Score threshold $\tau=0.1$} \\
    Mean NFR$_\tau$ ($|D_\tau(q_0)|$) $\downarrow$ & 0.607 & 0.224{\tiny$\pm$0.028} & 0.352 & 0.022{\tiny$\pm$0.004} & 1.263 & 0.563{\tiny$\pm$0.084} & 0.000 & 0.000{\tiny$\pm$0.000} \\
    Max NFR$_\tau$ ($|D_\tau(q_0)|$) $\downarrow$ & 1.003 & 0.435{\tiny$\pm$0.074} & 0.480 & 0.026{\tiny$\pm$0.006} & 1.921 & 0.747{\tiny$\pm$0.213} & 0.000 & 0.000{\tiny$\pm$0.000} \\
    Mean PFR$_\tau$ ($N-|D_\tau(q_0)|$) $\uparrow$ & 6.470 & 12.176{\tiny$\pm$1.557} & 55.701 & 78.434{\tiny$\pm$4.052} & 10.161 & 9.391{\tiny$\pm$0.929} & 0.000 & 0.000{\tiny$\pm$0.000} \\
    Mean SR$_\tau$ ($N$) $\downarrow$ & 0.206 & 0.017{\tiny$\pm$0.003} & 0.069 & 0.002{\tiny$\pm$0.001} & 0.190 & 0.003{\tiny$\pm$0.003} & 0.000 & 0.000{\tiny$\pm$0.000} \\
    Max SR$_\tau$ ($N$) $\downarrow$ & 0.732 & 0.059{\tiny$\pm$0.007} & 0.163 & 0.006{\tiny$\pm$0.003} & 0.378 & 0.013{\tiny$\pm$0.007} & 0.000 & 0.000{\tiny$\pm$0.000} \\
    \midrule
    \multicolumn{9}{@{}l}{Score threshold $\tau=0.25$} \\
    Mean NFR$_\tau$ ($|D_\tau(q_0)|$) $\downarrow$ & 1.388 & 0.373{\tiny$\pm$0.020} & 0.921 & 0.149{\tiny$\pm$0.005} & 3.081 & 0.773{\tiny$\pm$0.141} & 0.000 & 0.000{\tiny$\pm$0.000} \\
    Max NFR$_\tau$ ($|D_\tau(q_0)|$) $\downarrow$ & 2.324 & 0.815{\tiny$\pm$0.083} & 1.481 & 0.233{\tiny$\pm$0.017} & 6.817 & 1.018{\tiny$\pm$0.271} & 0.000 & 0.002{\tiny$\pm$0.003} \\
    Mean PFR$_\tau$ ($N-|D_\tau(q_0)|$) $\uparrow$ & 6.595 & 10.352{\tiny$\pm$0.277} & 45.731 & 69.642{\tiny$\pm$1.425} & 7.338 & 5.420{\tiny$\pm$0.646} & 0.000 & 0.000{\tiny$\pm$0.000} \\
    Mean SR$_\tau$ ($N$) $\downarrow$ & 0.343 & 0.065{\tiny$\pm$0.005} & 0.265 & 0.021{\tiny$\pm$0.008} & 0.500 & 0.022{\tiny$\pm$0.004} & 0.000 & 0.000{\tiny$\pm$0.000} \\
    Max SR$_\tau$ ($N$) $\downarrow$ & 1.254 & 0.203{\tiny$\pm$0.022} & 0.790 & 0.062{\tiny$\pm$0.021} & 1.479 & 0.067{\tiny$\pm$0.014} & 0.000 & 0.000{\tiny$\pm$0.000} \\
    \midrule
    \multicolumn{9}{@{}l}{Score threshold $\tau=0.5$} \\
    Mean NFR$_\tau$ ($|D_\tau(q_0)|$) $\downarrow$ & 3.295 & 1.136{\tiny$\pm$0.037} & 2.455 & 0.473{\tiny$\pm$0.062} & 11.367 & 1.674{\tiny$\pm$0.080} & n/a$^{\ddag}$ & 0.000{\tiny$\pm$0.001} \\
    Max NFR$_\tau$ ($|D_\tau(q_0)|$) $\downarrow$ & 4.683 & 2.300{\tiny$\pm$0.069} & 3.729 & 0.756{\tiny$\pm$0.089} & 36.002 & 2.304{\tiny$\pm$0.371} & n/a$^{\ddag}$ & 0.002{\tiny$\pm$0.003} \\
    Mean PFR$_\tau$ ($N-|D_\tau(q_0)|$) $\uparrow$ & 4.244 & 5.107{\tiny$\pm$0.288} & 47.021 & 57.732{\tiny$\pm$0.738} & 9.335 & 5.669{\tiny$\pm$0.356} & 0.000 & 0.000{\tiny$\pm$0.000} \\
    Mean SR$_\tau$ ($N$) $\downarrow$ & 0.112 & 0.041{\tiny$\pm$0.009} & 0.246 & 0.114{\tiny$\pm$0.022} & 0.038 & 0.014{\tiny$\pm$0.005} & 0.000 & 0.000{\tiny$\pm$0.000} \\
    Max SR$_\tau$ ($N$) $\downarrow$ & 0.431 & 0.147{\tiny$\pm$0.039} & 0.871 & 0.290{\tiny$\pm$0.050} & 0.091 & 0.045{\tiny$\pm$0.015} & 0.000 & 0.000{\tiny$\pm$0.000} \\
    \bottomrule
  \end{tabular*}

\end{table}

\FloatBarrier

\section{Sensitivity to Target Renaming}
\label{app:renaming}

\suppressfloats[t]
Table~\ref{tab:renaming} reports the complete target-renaming results, covering zero-shot and adapted models and both AP and AR; Section~\ref{sec:exp:renaming} summarizes the adapted-model AP. Within the full-category query, \emph{person} is replaced by \texttt{human}, \texttt{people}, \texttt{hiker}, \texttt{missing person}, \texttt{victim}, or \texttt{body}, while \emph{clothes} is replaced by \texttt{clothing}, \texttt{garment}, or \texttt{jacket}; the alternatives include near-synonyms as well as operationally plausible but semantically narrower or shifted descriptors.

Beyond the trends summarized in the main text, the full results show that semantic similarity between target names does not necessarily yield consistent detection performance. Differences between semantically related formulations are already present in the zero-shot models, indicating that pretrained vision--language representations do not guarantee invariance to target wording, and adaptation does not uniformly reduce these differences: several scenario-specific substitutes fall farther behind the canonical name after adaptation than before, and combinations such as \texttt{garment} for OmDet-Turbo diverge further from the canonical formulation. Thus, adaptation can improve detection accuracy without improving consistency across semantically related query formulations.

\begin{table}[!htp]
  \caption{Target renaming inside the full-category query.
  Each row replaces \emph{person} or \emph{clothes} with an alternative name
  and scores predictions against the original annotation class. Fine-tuned
  entries summarize three adaptation seeds.}
  \label{tab:renaming}

  \centering
  \footnotesize
  \setlength{\tabcolsep}{2.2pt}
  \begin{tabular*}{0.95\textwidth}{@{\extracolsep{\fill}}l cc cc cc cc@{}}
    \toprule
    & \multicolumn{2}{c}{GroundingDINO}
    & \multicolumn{2}{c}{GLIP}
    & \multicolumn{2}{c}{OmDet-Turbo}
    & \multicolumn{2}{c}{OWLv2} \\
    \cmidrule(lr){2-3} \cmidrule(lr){4-5}
    \cmidrule(lr){6-7} \cmidrule(lr){8-9}
    Text query & AP & AR & AP & AR & AP & AR & AP & AR \\
    \midrule
    \multicolumn{9}{@{}l}{\emph{Zero-shot} substitutes for \texttt{person}, \emph{person} AP and AR} \\
    \texttt{person} \emph{(reference)} & 57.80 & 71.36 & 56.79 & 70.23 & 54.29 & 65.93 & 58.00 & 74.39 \\
    \texttt{human} & 55.71 & 71.26 & 56.42 & 69.79 & 52.63 & 65.71 & 56.94 & 74.05 \\
    \texttt{people} & 59.04 & 74.19 & 57.15 & 70.53 & 53.03 & 65.09 & 56.83 & 74.21 \\
    \texttt{hiker} & 58.59 & 73.59 & 22.17 & 59.88 & 52.60 & 66.24 & 56.51 & 73.96 \\
    \texttt{missing person} & 53.73 & 74.52 & 55.98 & 70.19 & 51.64 & 63.92 & 55.65 & 73.89 \\
    \texttt{victim} & 58.93 & 73.97 & 49.81 & 65.78 & 47.09 & 65.59 & 54.52 & 73.60 \\
    \texttt{body} & 54.47 & 70.67 & 51.94 & 67.23 & 20.10 & 60.66 & 54.94 & 73.48 \\
    \midrule
    \multicolumn{9}{@{}l}{\emph{Zero-shot} substitutes for \texttt{clothes}, \emph{clothes} AP and AR} \\
    \texttt{clothes} \emph{(reference)} & 4.12 & 78.90 & 11.97 & 71.72 & 33.61 & 72.55 & 24.37 & 79.00 \\
    \texttt{clothing} & 3.81 & 78.64 & 12.68 & 72.26 & 31.82 & 72.39 & 15.27 & 78.88 \\
    \texttt{garment} & 31.34 & 76.50 & 18.44 & 69.33 & 33.71 & 72.74 & 22.06 & 79.24 \\
    \texttt{jacket} & 2.58 & 73.53 & 4.78 & 72.21 & 9.29 & 72.45 & 4.52 & 77.34 \\
    \midrule
    \multicolumn{9}{@{}l}{\emph{Fine-tuned} substitutes for \texttt{person}, \emph{person} AP and AR} \\
    \texttt{person} \emph{(reference)} & 64.30{\tiny$\pm$0.24} & 78.14{\tiny$\pm$0.19} & 67.93{\tiny$\pm$0.24} & 76.52{\tiny$\pm$0.12} & 55.26{\tiny$\pm$0.37} & 66.75{\tiny$\pm$0.16} & 59.00{\tiny$\pm$3.17} & 73.22{\tiny$\pm$1.02} \\
    \texttt{human} & 59.80{\tiny$\pm$2.15} & 75.10{\tiny$\pm$0.77} & 66.55{\tiny$\pm$0.81} & 76.21{\tiny$\pm$0.48} & 55.75{\tiny$\pm$0.43} & 66.58{\tiny$\pm$0.28} & 59.21{\tiny$\pm$2.91} & 73.19{\tiny$\pm$1.10} \\
    \texttt{people} & 63.44{\tiny$\pm$0.34} & 78.02{\tiny$\pm$0.10} & 67.57{\tiny$\pm$0.38} & 76.25{\tiny$\pm$0.26} & 55.28{\tiny$\pm$0.57} & 66.39{\tiny$\pm$0.42} & 58.39{\tiny$\pm$3.35} & 73.16{\tiny$\pm$1.08} \\
    \texttt{hiker} & 61.96{\tiny$\pm$1.82} & 76.83{\tiny$\pm$0.87} & 20.67{\tiny$\pm$6.48} & 71.03{\tiny$\pm$0.96} & 52.34{\tiny$\pm$0.31} & 66.05{\tiny$\pm$0.20} & 44.73{\tiny$\pm$5.56} & 72.34{\tiny$\pm$1.46} \\
    \texttt{missing person} & 39.27{\tiny$\pm$3.44} & 76.53{\tiny$\pm$0.48} & 64.84{\tiny$\pm$0.49} & 75.10{\tiny$\pm$0.23} & 56.36{\tiny$\pm$0.43} & 66.53{\tiny$\pm$0.18} & 57.63{\tiny$\pm$3.08} & 73.02{\tiny$\pm$1.13} \\
    \texttt{victim} & 54.35{\tiny$\pm$5.20} & 75.98{\tiny$\pm$0.68} & 46.22{\tiny$\pm$0.58} & 74.16{\tiny$\pm$0.57} & 54.25{\tiny$\pm$0.16} & 67.02{\tiny$\pm$0.09} & 55.37{\tiny$\pm$3.07} & 72.17{\tiny$\pm$1.66} \\
    \texttt{body} & 42.07{\tiny$\pm$7.63} & 74.48{\tiny$\pm$0.53} & 57.24{\tiny$\pm$3.76} & 74.43{\tiny$\pm$0.35} & 32.23{\tiny$\pm$1.82} & 66.11{\tiny$\pm$0.10} & 40.02{\tiny$\pm$3.04} & 72.54{\tiny$\pm$1.24} \\
    \midrule
    \multicolumn{9}{@{}l}{\emph{Fine-tuned} substitutes for \texttt{clothes}, \emph{clothes} AP and AR} \\
    \texttt{clothes} \emph{(reference)} & 68.97{\tiny$\pm$0.29} & 83.67{\tiny$\pm$0.13} & 70.26{\tiny$\pm$0.14} & 83.21{\tiny$\pm$0.06} & 56.02{\tiny$\pm$0.28} & 74.70{\tiny$\pm$0.15} & 64.74{\tiny$\pm$1.35} & 79.45{\tiny$\pm$0.77} \\
    \texttt{clothing} & 68.40{\tiny$\pm$0.29} & 83.54{\tiny$\pm$0.13} & 69.88{\tiny$\pm$0.27} & 83.31{\tiny$\pm$0.02} & 56.34{\tiny$\pm$0.20} & 74.51{\tiny$\pm$0.11} & 64.33{\tiny$\pm$1.33} & 79.36{\tiny$\pm$0.77} \\
    \texttt{garment} & 66.87{\tiny$\pm$0.55} & 83.52{\tiny$\pm$0.20} & 69.04{\tiny$\pm$0.18} & 83.20{\tiny$\pm$0.04} & 51.18{\tiny$\pm$0.80} & 74.15{\tiny$\pm$0.15} & 62.82{\tiny$\pm$2.18} & 79.16{\tiny$\pm$0.83} \\
    \texttt{jacket} & 59.26{\tiny$\pm$0.86} & 82.82{\tiny$\pm$0.07} & 67.13{\tiny$\pm$0.24} & 83.07{\tiny$\pm$0.11} & 41.91{\tiny$\pm$2.92} & 74.33{\tiny$\pm$0.18} & 45.20{\tiny$\pm$2.29} & 76.46{\tiny$\pm$0.75} \\
    \bottomrule
  \end{tabular*}

\end{table}

\FloatBarrier

\section{Person False Positives in Person-Absent Scenes}
\label{app:person-absent}

\begin{figure}[!tbp]
  \centering
  \includegraphics[width=\textwidth]{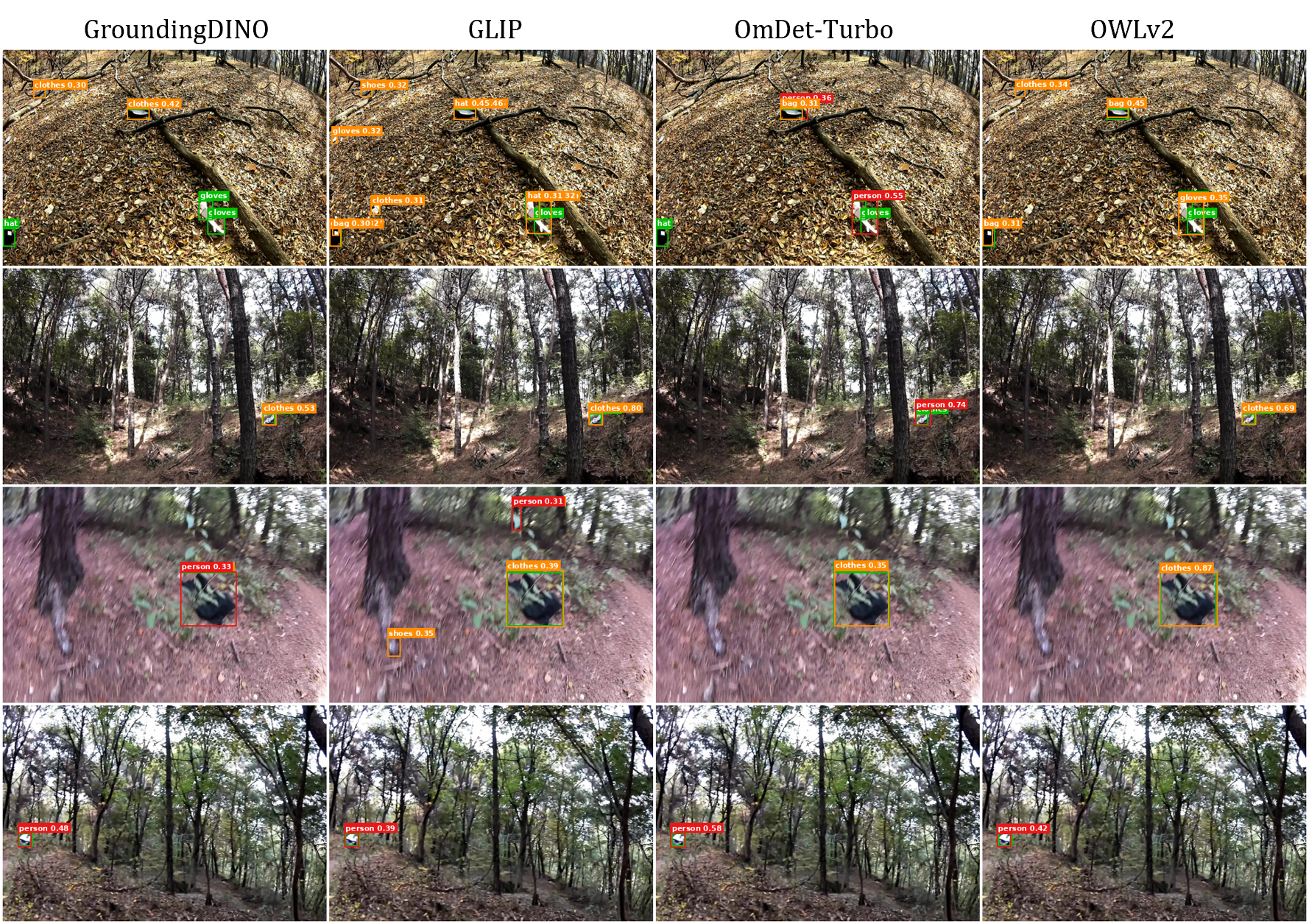}
  \caption{\textbf{Belongings misrecognition under the full-category query.}
  Predictions on person-absent scenes
  (red: \emph{person} predictions; orange: belongings predictions;
  green: annotated class).}
  \label{fig:belongings-confusions}
\end{figure}

The preceding contrasts are scored against person ground truth and therefore do
not capture a complementary failure mode: \emph{person} false positives in
scenes containing no person (Figure~\ref{fig:belongings-confusions}). Under the
assumed deployment setting, such detections constitute false alarms. A
\emph{person} alarm may prompt unnecessary inspection or responder deployment
and thereby consume limited operational resources. The following analysis
therefore quantifies \emph{person} false positives in person-absent scenes.

\begin{figure}[!tbp]
  \centering
  \includegraphics[width=\textwidth]{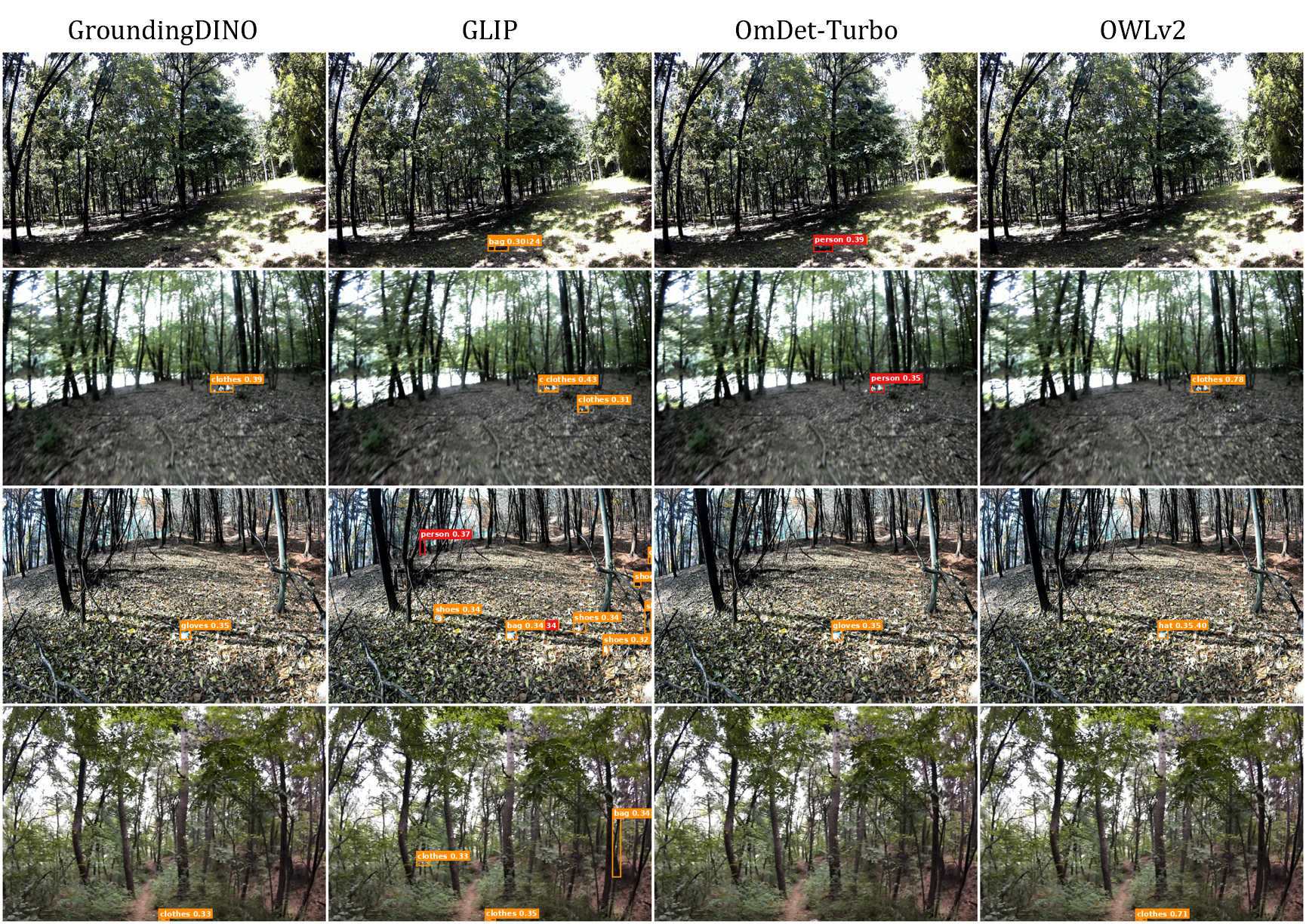}
  \caption{\textbf{False positives on negative samples containing neither a
  person nor belongings.} Predictions under the full-category query
  (red: \emph{person}; orange: belongings).}
  \label{fig:negative-frames}
\end{figure}

\begin{figure}[!tbp]
  \centering
  \includegraphics[width=\textwidth]{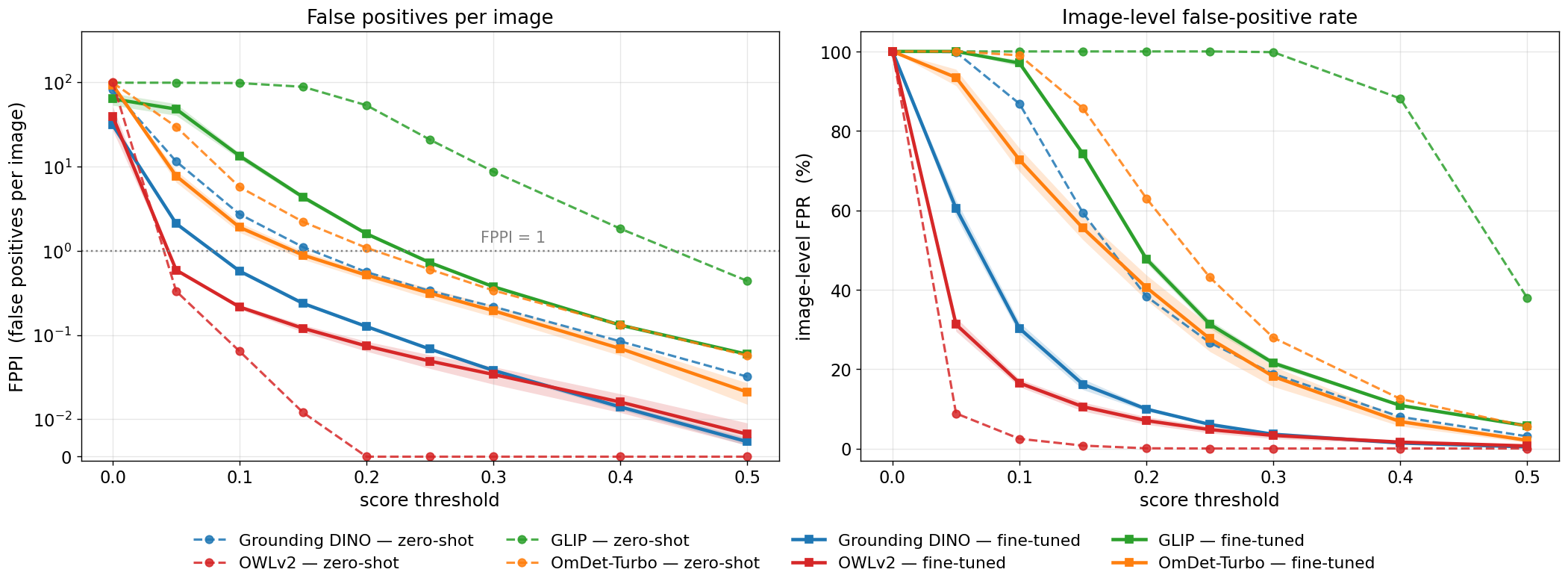}
  \caption{\textbf{Person false positives across score thresholds.} False
  positives per image (left, log scale) and the percentage of images
  with a false positive (right) under the full-category query. Dashed lines
  denote zero-shot models; solid lines and bands denote the fine-tuned mean
  and sample standard deviation over three seeds.}
  \label{fig:person-fp-sweep}
\end{figure}

The person-absent \dataset test set comprises $4{,}220$ frames containing
annotated belongings but no person and $180$ negative samples containing neither
an annotated person nor an annotated belonging, for a total of $4{,}400$ images.
Detectors can still produce \emph{person} false positives on the negative samples
(Figure~\ref{fig:negative-frames}), including high-scoring detections on small
forest-floor objects. Figure~\ref{fig:person-fp-sweep} reports
\emph{person} false positives over all $4{,}400$ images under the
full-category query across the full range of score thresholds. The reported
metrics are the number of false positives per image and the percentage of
images containing at least one false positive. At a score threshold of $0.0$,
the counts are dominated by the COCO limit of $100$ detections per image shared
across the queried categories. The resulting counts therefore reflect the
proposal budget rather than false-positive behavior; the analysis below
therefore reads the curves at the operating thresholds of $0.25$ and $0.5$.

The threshold sweep reveals substantial variation in zero-shot
false-positive behavior across detectors. At a score threshold of $0.25$, GLIP
produces approximately $21$ false positives per image, with false detections in
every image. OmDet-Turbo and GroundingDINO produce false positives in $43\%$ and
$27\%$ of the images, respectively, whereas OWLv2 produces no detections above
the threshold. Adaptation substantially reduces the false-positive rates of
the fusion-based detectors, with the largest reduction observed for GLIP.
Adapted OWLv2 instead shifts from zero false positives to a small but nonzero
rate. The relative ordering remains largely unchanged at the stricter
threshold, with GLIP continuing to exhibit the highest false-positive rate.
The results identify false-positive control as an important deployment
criterion alongside AP and AR because detectors with similar aggregate
accuracy can impose substantially different false-alarm burdens in
search and rescue operations.

Table~\ref{tab:person-absent-queries} compares the \texttt{person}-only query
with the full-category query on the $4{,}400$ person-absent images at a
score threshold of $0.25$. \emph{On-object} denotes false positives that
overlap an annotated belongings box at IoU $\geq 0.5$, whereas
\emph{Background} denotes boxes with no overlap with any annotation; boxes
with sub-threshold overlap appear only in the total count. The on-object
percentage is computed within each seed, and fine-tuned entries report the
mean and sample standard deviation over three seeds.

\begin{table}[H]
  \caption{\emph{Person} false positives on the $4{,}400$ person-absent
  images at score threshold $0.25$, comparing the \texttt{person}-only and
  full-category queries. Results are reported for \textbf{pretrained zero-shot}
  and \textbf{LoRA fine-tuned} OVDs; $^{\ddag}$ indicates no detection above
  the threshold.}
  \label{tab:person-absent-queries}
  \centering
  \footnotesize
  \setlength{\tabcolsep}{2.5pt}
  \begin{tabular*}{0.95\textwidth}{@{\extracolsep{\fill}}l cc cc cc cc@{}}
    \toprule
    & \multicolumn{2}{c}{GroundingDINO} & \multicolumn{2}{c}{GLIP}
      & \multicolumn{2}{c}{OmDet-Turbo} & \multicolumn{2}{c}{OWLv2} \\
    \cmidrule(lr){2-3} \cmidrule(lr){4-5} \cmidrule(lr){6-7} \cmidrule(lr){8-9}
    Metric & ZS & FT & ZS & FT & ZS & FT & ZS & FT \\
    \midrule
    \multicolumn{9}{@{}l}{Text query: \texttt{person}} \\
    \addlinespace[1pt]
    False positives & 3{,}977 & 5{,}172{\tiny$\pm$379} & 89{,}942 & 27{,}913{\tiny$\pm$1{,}606} & 3{,}686 & 3{,}790{\tiny$\pm$343} & 0$^{\ddag}$ & 216{\tiny$\pm$38} \\
    On-object & 2{,}936 & 4{,}064{\tiny$\pm$184} & 5{,}236 & 5{,}137{\tiny$\pm$199} & 2{,}330 & 2{,}736{\tiny$\pm$248} & 0 & 162{\tiny$\pm$38} \\
    Background & 818 & 822{\tiny$\pm$181} & 64{,}146 & 13{,}473{\tiny$\pm$483} & 949 & 686{\tiny$\pm$76} & 0 & 41{\tiny$\pm$6} \\
    On-object (\%) & 73.82 & 78.69{\tiny$\pm$2.97} & 5.82 & 18.45{\tiny$\pm$1.44} & 63.21 & 72.20{\tiny$\pm$0.53} & n/a$^{\ddag}$ & 74.50{\tiny$\pm$5.49} \\
    \midrule
    \multicolumn{9}{@{}l}{Text query: full-category} \\
    \addlinespace[1pt]
    False positives & 1{,}467 & 301{\tiny$\pm$6} & 91{,}457 & 3{,}172{\tiny$\pm$71} & 2{,}642 & 1{,}375{\tiny$\pm$197} & 0$^{\ddag}$ & 216{\tiny$\pm$38} \\
    On-object & 983 & 172{\tiny$\pm$9} & 5{,}440 & 1{,}182{\tiny$\pm$83} & 1{,}157 & 953{\tiny$\pm$138} & 0 & 162{\tiny$\pm$38} \\
    Background & 420 & 71{\tiny$\pm$3} & 76{,}357 & 1{,}381{\tiny$\pm$132} & 1{,}131 & 241{\tiny$\pm$42} & 0 & 41{\tiny$\pm$6} \\
    On-object (\%) & 67.01 & 57.31{\tiny$\pm$2.40} & 5.95 & 37.32{\tiny$\pm$3.31} & 43.79 & 69.34{\tiny$\pm$2.43} & n/a$^{\ddag}$ & 74.50{\tiny$\pm$5.49} \\
    \bottomrule
  \end{tabular*}
\end{table}

Adding the belongings categories substantially reduces \emph{person} false
positives for several model--adaptation configurations, although the effect is
model-dependent. For the adapted models, the false-positive count decreases by
factors of approximately $17$, $9$, and $3$ for GroundingDINO, GLIP, and
OmDet-Turbo, respectively. In contrast, zero-shot GLIP shows no reduction, and
OWLv2 produces identical detections under both queries. The box-level analysis
further reveals distinct false-positive profiles. For GroundingDINO,
OmDet-Turbo, and adapted OWLv2, $63$--$79\%$ of \emph{person} false positives
under the \texttt{person}-only query overlap annotated belongings, whereas only
about $6\%$ of zero-shot GLIP false positives overlap an annotated belonging at
IoU $\geq 0.5$.

The observed reductions are consistent with, but do not establish, a
class-assignment hypothesis. Under the \texttt{person}-only query,
belongings-like visual evidence may be assigned to the \emph{person} category
when no belongings category is available, whereas expanding the query
vocabulary provides alternative labels for that evidence. This interpretation
is consistent with the large reductions in on-object false-positive counts for
GroundingDINO and OmDet-Turbo. Zero-shot GLIP shows no corresponding reduction
in total false positives. Most zero-shot GLIP false positives occur in
background regions: $64{,}146$ of $89{,}942$ under the \texttt{person}-only
query and $76{,}357$ of $91{,}457$ under the full-category query. OWLv2, in
contrast, is query-invariant under the query comparison.

The adapted GLIP results further suggest that reassignment alone is
insufficient to explain the effect. The total false-positive count decreases by
approximately $25{,}000$, whereas the on-object count decreases by approximately
$4{,}000$. The difference between the reductions suggests that joint query
encoding may also alter the underlying predictions. The adapted GLIP behavior
complements the stealing-rate analysis in Table~\ref{tab:query-turnover}: if
adding a belongings category shifts evidence away from the \emph{person}
category, the same query interaction can suppress a true person detection when
a person is present or remove a person false positive when no person is present.

\FloatBarrier

\section{Assets, Licensing, and Intended Use}
\label{app:assets}

\paragraph{Code release.}

The source code and instructions for reproducing the experiments are available
at \url{https://github.com/DeokyunKim/OVD-forest-search}.

\paragraph{Existing assets.}

The datasets and model families used in the present study are cited at first use in the
main text. The core libraries and applicable licenses are as follows:
ForestPersons is distributed under the CC BY-NC-SA 4.0 license; GLIP under the
MIT license; GroundingDINO, OmDet-Turbo, OWLv2, and PEFT under Apache-2.0; and
pycocotools under BSD-2-Clause.

\paragraph{Release terms for \dataset.}

\dataset is released under the CC BY-NC-SA 4.0 license, which permits
non-commercial reuse with attribution and share-alike obligations. The imagery
was collected by the research team and a contracted agency. People appearing
in the released images consented to the use of the imagery for AI research.
Visible faces are blurred, and no identity or demographic information is
recorded or released.

\paragraph{Societal impact and prohibited uses.}

\dataset and the present study are intended to support missing-person search under
forest canopy. Uses that could harm people are discouraged, including autonomous
lethal weapons, surveillance of individuals, and attempts to identify people
appearing in the imagery.

\clearpage
\section*{NeurIPS Paper Checklist}

\begin{enumerate}

\item {\bf Claims}
    \item[] Question: Do the main claims made in the abstract and introduction accurately reflect the paper's contributions and scope?
    \item[] Answer: \answerYes{}.
    \item[] Justification: The abstract and Section~\ref{sec:intro} state the claims and contributions.

\item {\bf Limitations}
    \item[] Question: Does the paper discuss the limitations of the work performed by the authors?
    \item[] Answer: \answerYes{}.
    \item[] Justification: Limitations are stated in Section~\ref{sec:exp:renaming}.

\item {\bf Theory assumptions and proofs}
    \item[] Question: For each theoretical result, does the paper provide the full set of assumptions and a complete (and correct) proof?
    \item[] Answer: \answerNA{}.
    \item[] Justification: The paper contains no theoretical results.

    \item {\bf Experimental result reproducibility}
    \item[] Question: Does the paper fully disclose all the information needed to reproduce the main experimental results of the paper to the extent that it affects the main claims and/or conclusions of the paper (regardless of whether the code and data are provided or not)?
    \item[] Answer: \answerYes{}.
    \item[] Justification: Sections~\ref{sec:method:queries} and~\ref{sec:exp:setup} and Appendices~\ref{app:dataset} and~\ref{app:setup-details} specify the protocol, training settings, queries, and splits. The  code repository linked in Appendix~\ref{app:assets} provides the source code and reproduction instructions.

\item {\bf Open access to data and code}
    \item[] Question: Does the paper provide open access to the data and code, with sufficient instructions to faithfully reproduce the main experimental results, as described in supplemental material?
    \item[] Answer: \answerYes{}.
    \item[] Justification: \dataset is available at the  URL in the abstract, and the source code and reproduction instructions are available through the  repository linked in Appendix~\ref{app:assets}.

\item {\bf Experimental setting/details}
    \item[] Question: Does the paper specify all the training and test details (e.g., data splits, hyperparameters, how they were chosen, type of optimizer) necessary to understand the results?
    \item[] Answer: \answerYes{}.
    \item[] Justification: Section~\ref{sec:exp:setup} reports the main training and evaluation settings; Appendix~\ref{app:setup-details} gives model-specific details and Appendix~\ref{app:dataset} gives the splits.

\item {\bf Experiment statistical significance}
    \item[] Question: Does the paper report error bars suitably and correctly defined or other appropriate information about the statistical significance of the experiments?
    \item[] Answer: \answerYes{}.
    \item[] Justification: Fine-tuned results are reported as mean $\pm$ sample standard deviation over three seeds (Section~\ref{sec:exp:setup}); zero-shot evaluation is a single deterministic run.

\item {\bf Experiments compute resources}
    \item[] Question: For each experiment, does the paper provide sufficient information on the computer resources (type of compute workers, memory, time of execution) needed to reproduce the experiments?
    \item[] Answer: \answerYes{}.
    \item[] Justification: GPU, per-model runtime, and total budget are in Section~\ref{sec:exp:setup} and Table~\ref{tab:compute}.
    
\item {\bf Code of ethics}
    \item[] Question: Does the research conducted in the paper conform, in every respect, with the NeurIPS Code of Ethics \url{https://neurips.cc/public/EthicsGuidelines}?
    \item[] Answer: \answerYes{}.
    \item[] Justification: The research conforms to the Code of Ethics.

\item {\bf Broader impacts}
    \item[] Question: Does the paper discuss both potential positive societal impacts and negative societal impacts of the work performed?
    \item[] Answer: \answerYes{}.
    \item[] Justification: Appendix~\ref{app:assets} discusses the intended benefit, misuse risks, privacy precautions, and licensing terms.
    
\item {\bf Safeguards}
    \item[] Question: Does the paper describe safeguards that have been put in place for responsible release of data or models that have a high risk for misuse (e.g., pre-trained language models, image generators, or scraped datasets)?
    \item[] Answer: \answerYes{}.
    \item[] Justification: Appendix~\ref{app:assets} describes the release safeguards: no trained model is released, \dataset is distributed under the non-commercial CC BY-NC-SA 4.0 license, visible faces are blurred and no identity or demographic information is recorded, and prohibited uses including surveillance and autonomous weapons are stated.

\item {\bf Licenses for existing assets}
    \item[] Question: Are the creators or original owners of assets (e.g., code, data, models), used in the paper, properly credited and are the license and terms of use explicitly mentioned and properly respected?
    \item[] Answer: \answerYes{}.
    \item[] Justification: The datasets and model families are credited, and the applicable asset and core-library licenses are listed in Appendix~\ref{app:assets}.

\item {\bf New assets}
    \item[] Question: Are new assets introduced in the paper well documented and is the documentation provided alongside the assets?
    \item[] Answer: \answerYes{}.
    \item[] Justification: \dataset is documented in Section~\ref{sec:method:benchmark}, Appendix~\ref{app:dataset}, and Appendix~\ref{app:assets}.

\item {\bf Crowdsourcing and research with human subjects}
    \item[] Question: For crowdsourcing experiments and research with human subjects, does the paper include the full text of instructions given to participants and screenshots, if applicable, as well as details about compensation (if any)? 
    \item[] Answer: \answerYes{}.
    \item[] Justification: \dataset does not target human subjects; agency staff captured incidentally gave consent and are de-identified by face blurring. Data were collected and annotated partly by an agency contracted by our institute and partly in-house; compensation followed the contract and local labor standards, and no crowdsourcing platforms were used.

\item {\bf Institutional review board (IRB) approvals or equivalent for research with human subjects}
    \item[] Question: Does the paper describe potential risks incurred by study participants, whether such risks were disclosed to the subjects, and whether Institutional Review Board (IRB) approvals (or an equivalent approval/review based on the requirements of your country or institution) were obtained?
    \item[] Answer: \answerNA{}.
    \item[] Justification: The dataset does not contain real missing persons or any sensitive personal or biometric information.

\item {\bf Declaration of LLM usage}
    \item[] Question: Does the paper describe the usage of LLMs if it is an important, original, or non-standard component of the core methods in this research? Note that if the LLM is used only for writing, editing, or formatting purposes and does \emph{not} impact the core methodology, scientific rigor, or originality of the research, declaration is not required.
    \item[] Answer: \answerNA{}.
    \item[] Justification: The research does not involve the use of large language models (LLMs).

\end{enumerate}

\end{document}